\documentclass[preprint,12pt]{elsarticle}

\usepackage{graphics}
\usepackage{amsthm} 
\usepackage{color}
\usepackage{setspace}
\usepackage{epsfig}
\usepackage{epstopdf}
\usepackage{graphicx}
\usepackage{url}    
\usepackage{float}
\usepackage{varwidth}
\usepackage{comment}
\usepackage{paralist}   
\usepackage[cmex10]{amsmath} 
\usepackage[ruled,vlined,commentsnumbered,linesnumbered]{algorithm2e}
\usepackage{algorithm2e,float}
\usepackage{amssymb}
\journal{Image and Vision Computing}

\begin{document}

\begin{frontmatter}


 \author{Smit Marvaniya\corref{cor1}}
 \ead{smit@cse.iitm.ac.in}
% \address{Indian Institute of Technology Madras, Chennai, INDIA. \fnref{label3}}
% \fntext[label3]{+918095579399}

\author{Raj Gupta}
\ead{gupta.raj@ieee.org}

\author{Anurag Mittal}
\ead{amittal@cse.iitm.ac.in}
\ead[url]{www.cse.iitm.ac.in/~amittal}
\address{Computer Science and Engineering Department, \\Indian Institute of Technology Madras, \\Chennai, INDIA - 600036. \fnref{label3}}

\title{Adaptive Locally Affine-Invariant Shape Matching}


\begin{abstract}
Matching deformable objects using their shapes is an important problem in computer vision since shape is perhaps the most 
distinguishable characteristic of an object.  The problem is difficult due to many factors such as intra-class variations, 
local deformations, articulations, viewpoint changes and missed and extraneous contour portions due to errors in shape
extraction.  While small local deformations has been handled in the literature by allowing some leeway
in the matching of individual contour points via methods such as Chamfer distance and Hausdorff distance, handling more severe deformations and
articulations has been done by applying local geometric corrections such as similarity or affine.
However, determining which portions of the shape should be used for the geometric corrections is very hard, although some methods have been
tried.  In this paper, we address this problem by an efficient 
search for the group of contour segments to be clustered together for a geometric correction using Dynamic Programming by essentially searching for the
segmentations of two shapes that lead to the best matching between them.  At the same time, we allow portions of the contours to remain 
unmatched to handle missing and extraneous contour portions.
Experiments indicate that our method outperforms other algorithms, especially when the shapes to be matched are more complex.
\end{abstract}


\end{frontmatter}


\section{Introduction}
\label{Intro}
Matching of deformable object shapes is an interesting as well as an important problem in Computer Vision 
since shape is one of the most distinguishing characteristics of an object, being unaffected by photometric changes 
and background variations.  Furthermore, it has been found from human perception that, in the presence of challenges 
such as partial occlusions, local articulations, geometric distortions, intra-class variations and viewpoint changes, 
it is possible to identify and recognize an object simply from its shape.  Thus, shape matching has been successfully 
used in various tasks such as Object Detection and Classification (\cite{GuoIJCV14,wangCVPR12,CG,MC}), Optical Character Recognition \cite{SC}, 
Medical Image Registration \cite{HuangPAMI2006} and Image Retrieval \cite{NPADI}.

\begin{figure}
\centering
\begin{tabular}{cc}
\includegraphics[height=4.0cm,width=6.5cm]{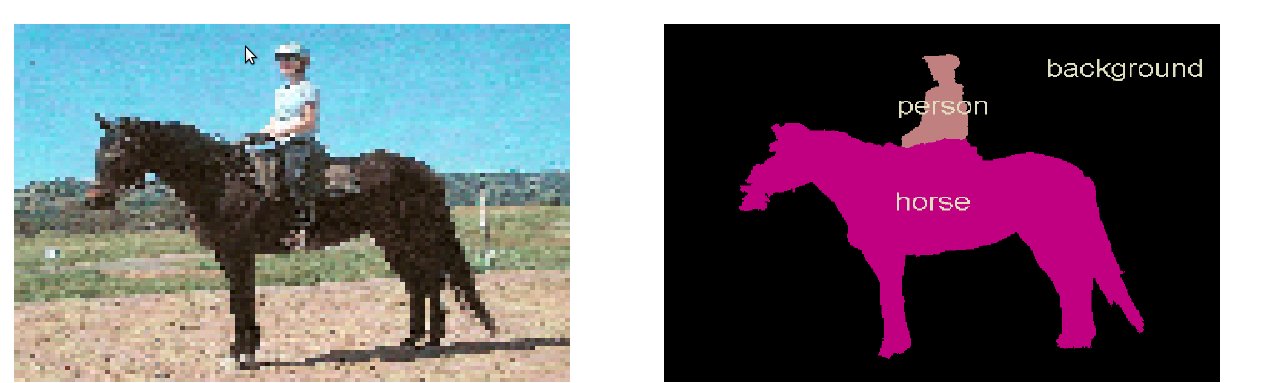}&
\includegraphics[height=4.0cm,width=6.5cm]{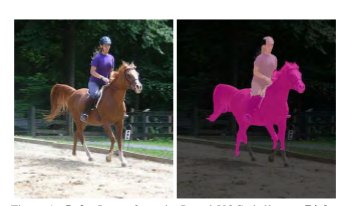} \\
(a) & (b)\\
\end{tabular}
 \caption {Some segmentations obtained from the standard Image Segmentation algorithms of Russell et al. \cite{RussellICCV2009} 
and Brox et al. \cite{BroxCVPR2011} respectively.} 
\label{Inputs_BG_IS}
\end{figure}

However, the task of modeling such variations as mentioned above in a computer is quite challenging.  
Furthermore, in real scenarios, when an object is segmented out automatically using techniques such 
as Background Subtraction or Image Segmentation, the matching of the extracted contours should be 
robust to errors introduced by the segmentation process.  For instance, the output of a Background Subtraction technique often misses out 
some portions of the object or adds some extra portions such as object shadows.  Sometimes,  
two objects may be merged into one if they are close to each other.  Similar problems exist due to the use of Image Segmentation techniques as well.  
Figure \ref{Inputs_BG_IS} shows the output obtained from standard Image Segmentation algorithms of 
Russell et al. \cite{RussellICCV2009} and Brox et al. \cite{BroxCVPR2011} respectively.  
Note that the legs of the horse are combined together in Figure \ref{Inputs_BG_IS}(a) whereas some of the leg portions 
of the horse are missed out in Figure \ref{Inputs_BG_IS}(b).  A robust shape matching algorithm must deal with such 
variations and distractions in order to be useful in a practical scenario.  

Several algorithms have been considered in the past for the problem of shape matching.
As in \cite{DZPR2004}, most of these can be broadly classified into two categories 
based on the types of features used: \begin{inparaenum} \item methods that treat the shape as a blob in order to come up with 
an approximate representation, and \item methods that use the contour boundary information directly.  \end{inparaenum}

One of the most popular blob-based approaches is to use a shock graph or a medial-axis transform for shape representation.
Techniques that use these (\cite{MacriniDCVIU2011,SiddiqiKIJCV1999,SebastianPAMI2004,Benjamin94,NAlajlanPAMI2008}), first 
build a graph that models the skeleton of a shape.  
Topological similarity between the graphs helps in identifying the global shape structure, whereas geometric similarity 
at every node helps to capture the local shape information.  These methods perform well in the presence of deformations.  
However, they build the skeleton \emph{a-priori} and can only match shapes when there is an overall global similarity between 
them and may fail in the presence of articulations, occlusions or noise in shape extraction.  
 
To deal with such challenges some methods (\cite{Felzenszwalb05,ABronsteinIJCV2008}) segment the shape into regions or parts that 
are then used for the task of matching. 
These methods capture the local shape variation much better by allowing articulations of such portions about each other.
Felzenszwalb \cite{Felzenszwalb05} proposed a technique for shape representation based on triangulated polygons and used 
 Dynamic Programming to match such representations.  However, this method can lead to errors in the presence of occlusions.  
To handle partial matching of shapes, Bronstein et al. \cite{ABronsteinIJCV2008} proposed a {\em pareto} framework to determine 
an optimal tradeoff between part similarity and part decomposition.  Although the ideas in this paper are quite close to our work,
the method is computationally expensive due to working on shape blobs.  Furthermore, the optimization method proposed to search 
over the parts is computationally very expensive and gives only an approximate solution. 

While it may be claimed that blob-based methods are more robust due to the consideration of the entire 2D space, such methods tend to be 
computationally very expensive due to the processing of all the pixels enclosed by a shape.  Thus, it is much more efficient to use the
boundary information alone  in order to match shapes. 
  
Methods that use only contour boundary information for shape representation can further be classified 
into \begin{inparaenum} \item Global \item Part-based methods. \end{inparaenum}  
  
Many global methods exist such as shape descriptors (\cite{SC,MoriPAMI2005,LateckiCVPR200}), shape distances (\cite{HD,FDCM}) 
and contour matching techniques (\cite{DaliriPR2008,NPADI,LVQ,YZIIS2011,Basri1998,LLateckiPAMI2000,BrynerPAMI2013}), some of which also
estimate the affine or projective transformation required to match the shapes \cite{SC,BrynerPAMI2013}.
Among these, Shape Context \cite{SC} is a popular method that builds a shape descriptor using the Euclidean distance 
and the relative orientation of the contour points in a log-polar space.  The dissimilarity between two shapes is a weighted sum of 
the matching errors, computed using a maximum bipartite algorithms and a measure on the transformation required to match the two shape contours.  
The method works well to match shapes invariant to rigid transformations and deals with small deformations present in the contour boundary.  
One of the popular extensions of Shape Context (proposed by Mori et al. \cite{MoriPAMI2005}) solves the problem of shape matching 
very efficiently using multi-stage pruning techniques.  
The first stage is called representative shape contexts that matches very few shape contexts and identifies the outliers very fast, 
whereas the second stage matches the shapes in more details based on vector quantization in the space of shape contexts 
that involves clustering of the vectors, called as shapemes. 
All these methods rely on global features and hence fail in the presence of articulations, partial occlusions and noise present in the contour boundary.

To address such problems,  Hong et al. \cite{HongCVPR2006} and Adamek and O'Connor \cite{AdamekCSVT2004} represent shapes in terms 
of local features such as concave or convex portions of a contour to preserve the local geometry.  Even though these more local methods 
are robust to some deformations, articulations and noise, they do not preserve sufficient contour information for a very discriminative matching.  
To model shapes better, the techniques mentioned in \cite{FPS} and \cite{XChunjingPAMI2009} combine local and global 
features.  Felzenszwalb and Schwartz \cite{FPS} proposed a hierarchical matching technique for deformable shapes 
even in the presence of a cluttered background wherein a tree is built whose leaf nodes capture the local information and nodes 
close to the root capture global information.  A Dynamic Programming based matching technique is used to match the two \emph{shape trees}.  
On the other hand, Xu et al. \cite{XChunjingPAMI2009} proposed a \emph{Contour Flexibility} descriptor that gives a deformation potential to 
each contour point so as to deal with deformations.  The similarity between shapes is calculated by considering a linear combination of the 
local and the global measures.  Although these methods use both local and global features and perform well against deformations,
they do not consider partial matching of shapes and so may fail in the presence of occlusions.  
  
In order to handle occlusions, Latecki et al. \cite{LVQ} developed an elastic shape matching algorithm based on an efficient Dynamic Programming based 
matching approach.  This method identifies the outliers that may be present in the query shapes by allowing skips while matching. 
However, this method fails to capture the part structure of a shape and so may perform poorly in the presence of articulations.

There have been numerous research efforts (\cite{GMcNeillCVPR2006,LFTCVPRW08,ATemlyakovCVPR2010,SBGI,YZIIS2011}) to deal 
with the local shape variations of an object shape and solve the problem of articulations.  
Cao et al. \cite{YZIIS2011} proposed an approach for matching shape contours using the ``procrustes'' distance 
between shapes \cite{dryden-mardia:98} and handles occlusions and shape segmentation by an MCMC (Markov chain Monte Carlo)-based 
search for the matching segments in two contours.
However, this method is computationally quite expensive due to the consideration of individual point-point matchings 
across the shapes and the MCMC iterations required for optimizing such point-point matchings for the whole shape.   
Ma et al. \cite{SBGI} proposed a technique for partial matching using geometric relations of shape
context as shape descriptor followed by maximal clique inference based hypothesis used to identify the best possible part correspondences.  
Although, this method handles the problem of partial occlusions, it may fail in the presence of articulations as the local descriptors are not 
restricted to capturing information only within a part.  Furthermore, the method is computationally very expensive due to the use of sampling
methods.

Another popular approach for handling articulations is the Inner Distance Shape Context (IDSC) proposed by 
Ling and Jacobs \cite {HJD} that solves the problem of articulation in certain scenarios 
and can be considered as an improvement over Shape Context \cite{SC}.  
This method builds a descriptor based on the relative spatial distribution of the contour points using the 
Inner Distance instead of the Euclidean distance, and the Inner Angle instead of the regular angle.  
The Inner Distance ($ID$) between a pair of contour points is defined as the length of the shortest path between them while totally 
remaining within the shape and the Inner Angle is the angle from one point to the other that is in the direction of this 
shortest ``inner'' path.  A Dynamic Programming-based algorithm instead of a bipartite matching approach was also introduced by taking 
advantage of the ordering constraint in order to solve the point correspondence problem.  
This method is invariant to the 2D-articulations of a shape as it captures the part structure effectively.
However, it is not invariant to affine changes of individual parts and also fails under partial occlusions as all the contour 
points are considered while building the descriptor and while matching. 

In order to handle local affine changes, Gopalan et al. \cite{RPRC} proposed a shape-decomposition technique 
that divides a shape into convex parts using Normalized Cuts \cite{JJM}.  These parts are then individually affine normalized 
and combined into a single shape that is matched using IDSC.  As a result, this method is able to capture more deformations of local portions, 
such as a 3D part articulation that may be modeled by a 2D affine transformation of its projection.  
This yields a significant improvement over IDSC in many cases.  It nevertheless assumes an \emph{a-priori} shape decomposition 
from a single shape that may be inconsistent in the presence of occlusions or noise in shape extraction. 
Furthermore, the matching is still global and hence one will be unable to handle partial occlusions of the shapes.

In this work, we propose a locally deformable matching technique that 
does not require one to make an \emph{a priori} assumption about the decomposition of a shape contour.
Rather, the contour decomposition is determined during matching by an efficient search for the decompositions of two contours (into Groups-of-Segments (GSs))
that yield the best matching.  The technique not only handles articulations, but also models occlusions and extraneous segments  
explicitly by skipping non-matching segments 
during matching.
Furthermore, each such Group-of-Segments (GSs) is affine-corrected before matching which uses a robust contour matching technique 
that handles deformations well.
As a result, our method is robust in the presence of various challenges such as intra-class variations, articulations, deformations, partial occlusions
and errors in the shape extraction process.  This is illustrated by results that show significant improvement over the state-of-the-art, 
especially in the case of partial occlusions and errors in shape extraction.

The remainder of the paper is organized as follows: Section 2 describes the processes of shape representation and
the extraction of possible GSs from shapes.  The cost function for shape similarity given a particular GS 
correspondence across shapes is described in Section 3.  Section 4 describes the process of efficiently determining the best GS correspondence
that minimizes this cost function using Dynamic Programming.  Finally, in Section 5, we show some promising results obtained by our method.

%%%%%%%%%%%%%%%%%%%%%%%%%%%%%%%%%%%%%%%%%%%%%%%%%%%%%%%%%%%%%%%%%%%%%%%%%%%%%%%%%%%%%%%%%%%%%%%%%%%%%%%%%%%%%
%----------------------------------SECTION  2--Shape Representation-----------------------------------------%
%%%%%%%%%%%%%%%%%%%%%%%%%%%%%%%%%%%%%%%%%%%%%%%%%%%%%%%%%%%%%%%%%%%%%%%%%%%%%%%%%%%%%%%%%%%%%%%%%%%%%%%%%%%%%
\section{Shape Representation}
The first task for any shape matching technique is to come up with a representation of shapes 
such that they can be matched efficiently and accurately.  In our problem, the 
inputs are assumed to be outer shape contours, that may be obtained from automatic techniques such as background 
subtraction and Image Segmentation or in some cases, they may be manually drawn. 

\begin{figure*}
  \centering
\begin{tabular}{cc}
  \includegraphics[scale=0.3]{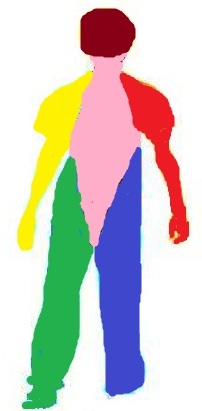}	&
  \includegraphics[scale=0.38]{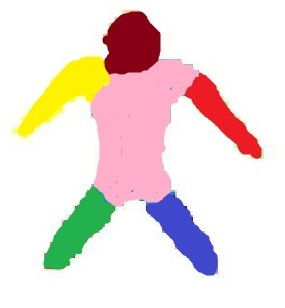}	\\
(a) & (b)
\end{tabular}
   \caption{Best viewed in color.  Part-decomposition obtained from Gopalan et al. \cite{RPRC} using a single shape above.} 
\label{Part_Decomposition_single_image}
\end{figure*}

Shape Contours have often been represented by decomposing them into small parts so that local transformations can be 
determined for each part.  
%A part-based model is used in this work as well since it can handle the articulations between the different parts of a shape better.
While decomposition of a shape is important, 
it is, however, a very challenging task, especially under occlusions or noise.  For example, results of shape-decomposition 
using a single shape using the method proposed by Gopalan et al. \cite{RPRC} are shown in Figure 
\ref{Part_Decomposition_single_image} and it may be inferred that the method more or less fails in consistently segmenting a
shape into the same parts in different shape instances.  This lead to errors when the individual parts across such shapes are 
attempted to be matched.  To deal with this problem, in this work, we consider multiple decomposition possibilities in our 
shape representation and chose the shape decomposition pair that matches best across two given shapes.  

First, we break the whole shape contour into small straight line-like segments.  Then, Groups of such Segments (GSs)
are created.  Then, the shape decompositions that are allowed for a given shape are taken to be collections of such
GSs such that they don't overlap, with some skips/unassigned segments allowed in the shape decomposition.
We next discuss how to extract possible \emph{break-points} from a shape contour that will form the 
possible start and end points of GSs.  This is done in order to restrict the number of points at which the GSs can 
start and end.

\subsection{Computing Possible Break Points for GSs}
Typically, points at which a local geometric transformation changes are coincident with points of high curvature.  
Hence, the points of high curvature are determined first.  Several approaches 
\cite{FMokhtarianPAMI1998,XCHeICPR2004,HeChenOE2008,TTuytelaarsECCV2008,LiuIJCV2008} have been proposed 
in the literature for high curvature point detection on contours.  
In this work, we try to detect an optimal number of points that are distributed across the entire contour and are
also robust to possible noise in the contour.  To this end, we first calculate \emph{Angle Sharpness} $S_a$ 
at a contour point $q_i$ using $NL$ (Neighborhood point list) which contains $N_{s}$ points on either side of $q_i$.
\emph{Angle Sharpness} $S_{a}$ for a contour point $q_i$ is defined as:
\begin{equation}
S_{a}(q_i) = \sum_{(q_{i-j},q_{i+j}) \in NL_{q_i}} w_j\cdot(180 - A(q_{i-j}, q_i, q_{i+j})),
\end{equation}
\begin{figure*}
  \centering
\begin{tabular}{cc}
  \includegraphics[scale=0.84]{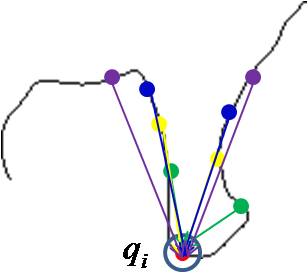} &
  \includegraphics[scale = 0.2]{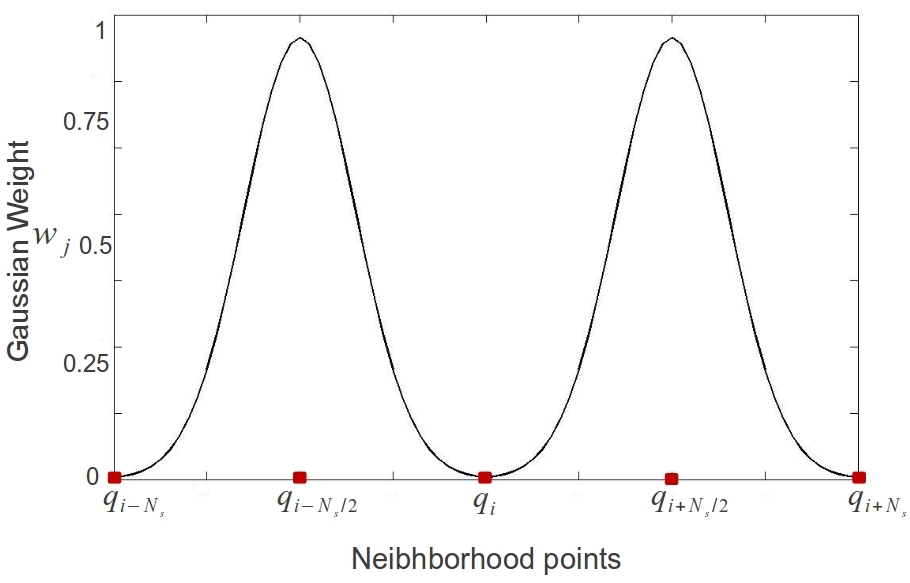} \\
(a) & (b)
\end{tabular}
  \caption {(a) Some of the possible neighborhood points around the contour point $q_i$ to calculate \emph{Angle Sharpness}.  
Pairs of the neighborhood points are shown in the same color. (b) The Gaussian weighting function.}
\label{Neighborhood_points}
\end{figure*}
where $A(q_{i-j}, q_i, q_{i+j})$ is the angle between the 
contour points $q_{i-j}$, $q_i$ and $q_{i+j}$.  To calculate the curvature at a certain scale which depends on $N_s$, we introduce 
a scheme whereby two Gaussians centered at $q_{i+(N_{s}/2)}$ and $q_{i-(N_{s}/2)}$ respectively are used.  The Gaussian weighting 
function makes the whole procedure quite robust to noise since the result depends on many points and not on a single point alone.
Figure \ref{Neighborhood_points} shows an example of such a computation.

Candidate curvature points are identified by considering local maxima of the \emph{Angle Sharpness} above a certain
threshold.  Furthermore, lower maxima close to a higher maximum are removed as they provide more or less duplicate 
information.  We call the points thus detected as \emph{high curvature points}.

\begin{figure}
\centering
\includegraphics[scale =0.35]{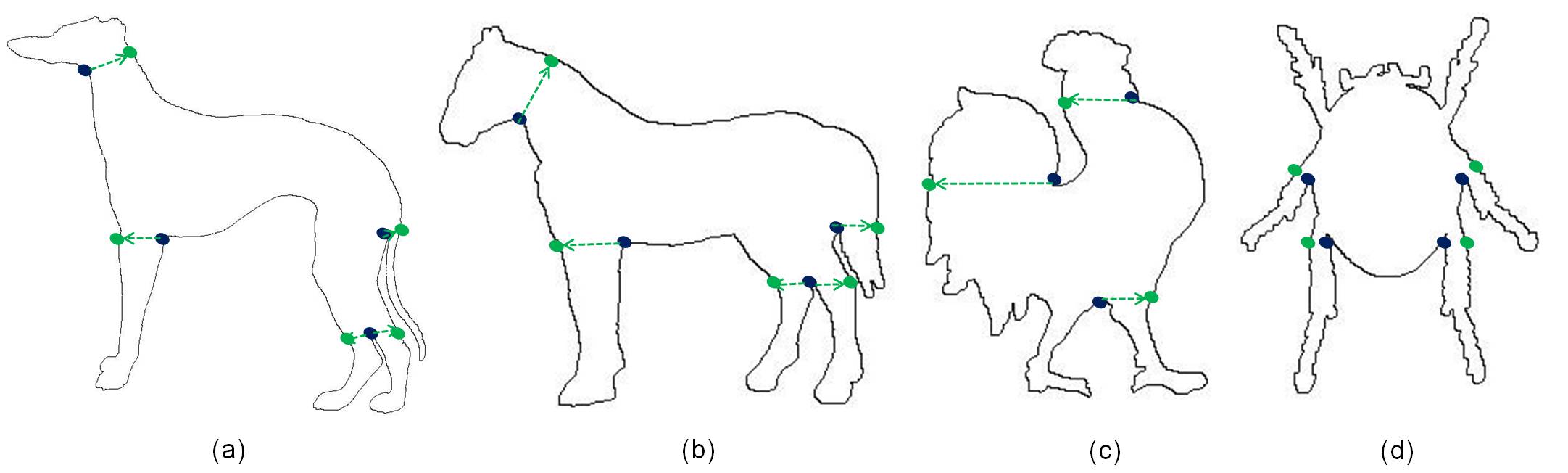}
 \caption {Possible non-convex articulated points (\emph{opposite points}) extracted by our algorithm.} 
\label{breakpoints}
\end{figure}

We further notice that all possible break points between local groups of segments cannot be modeled using only high curvature points as GS junctions.  
Figure \ref{breakpoints} shows points (in green) across which articulation occurs and 
the local transformation of the object shape often changes.  
Thus, we detect additional points known as \emph{opposite points} in this work.  

\begin{figure*}
  \centering
\begin{tabular}{cc}
  \includegraphics[scale=0.25]{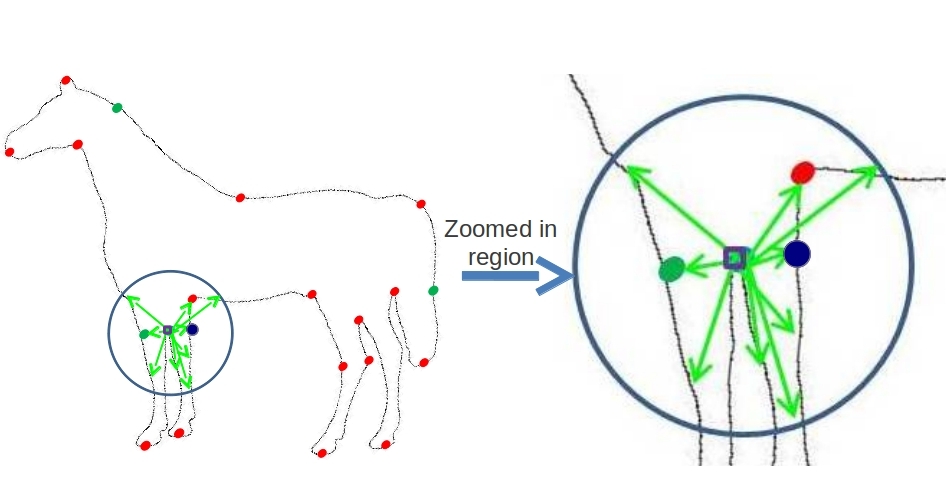}	&
  \includegraphics[scale=0.25]{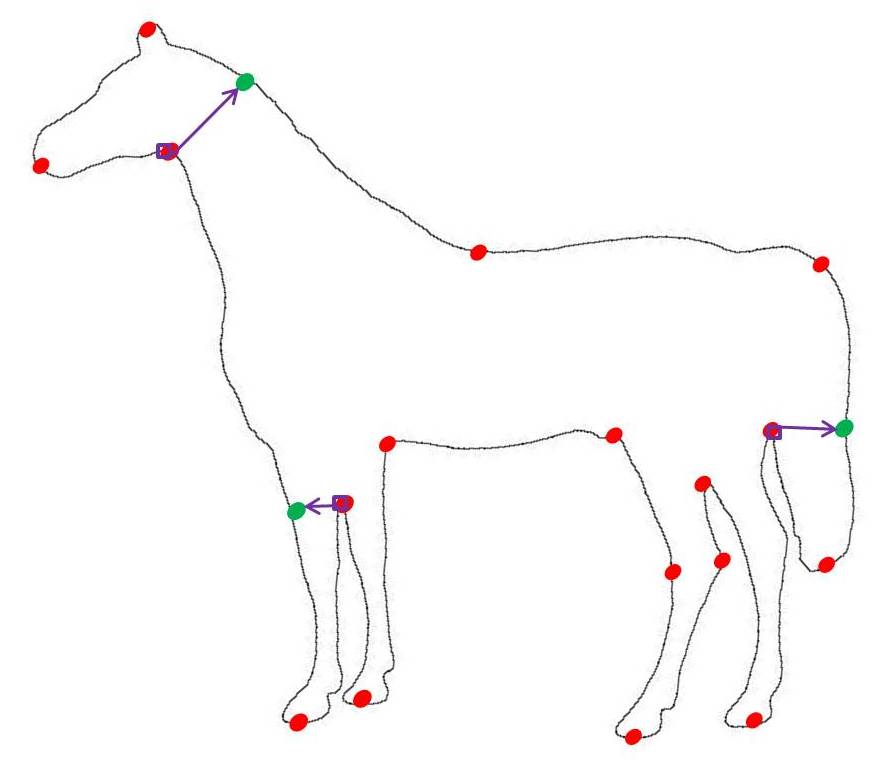}	\\
(a) & (b)
\end{tabular}
   \caption{(a) An example of computing the \emph{opposite point} for a given \emph{concave point}. (b) \emph{Opposite points} determined for a 
horse shape contour.} 
\label{Opposite-point-extraction}
\end{figure*}

To detect an \emph{opposite point}, we first determine \emph{concave points} based on the complexity measure proposed by 
Gopalan et al. \cite{RPRC}.  It has been observed in several prior works (\cite{LiuCVPR2010,Hoffman1984,EladPAMI2003,RPRC}) that each part of a shape is typically convex.   
Any two points within a convex region have the same Euclidean Distance ($ED$) 
and Inner Distance ($ID$) while a concave region has different $ID$ and $ED$ where the inner-distance ($ID$)\cite{HJD} is defined as the 
length of the shortest path within the shape boundary.  The shortest path is a collection of line segments and the intermediate vertice(s) 
on such shortest paths between points lying in two different convex regions
represent the \emph{concave points}.  Although these \emph{concave points} typically coincide with high 
curvature points detected in our approach, they help in extracting \emph{opposite points} since they are at the joint 
of two convex parts.   Figure \ref{Opposite-point-extraction} shows such a \emph{concave point} with a square.

For a given \emph{concave point} $p$, we consider all the contour points at Geodesic distance less than or equal to  a 
distance $d$ as candidate \emph{opposite points}.  This distance $d$ is fixed as $20$ percent of the total number of contour points in 
the contour.  The candidate \emph{opposite point} $c$, for which the distance between $p$ and $c$ is a local minimum and 
the ED and geodesic distance ($GD$) is sufficiently large, is considered as the \emph{opposite point} for \emph{concave point} $p$.
The green point in Figure \ref{Opposite-point-extraction}(a) is such an \emph{opposite point}.  
If there is any high curvature point in a close neighborhood of $c$, then $c$ represents the same information 
as that point and it is therefore taken to be the \emph{opposite point} instead of point $c$. 
This is again demonstrated in Figure \ref{Opposite-point-extraction}(a) using a blue \emph{opposite point} and red \emph{high curvature point}. 
The procedure for extracting the \emph{opposite point} is very similar to \cite{Mittal:detector12}.
Both \emph{opposite points} and \emph{high curvature points} are considered as possible \emph{break-points} in this work.  

There are cases where either the shapes or its portions are simple and hence, the possible break points in them cannot be determined, 
especially from a single shape. For e.g. if the
hand of a person is straight, it is very hard to determine from a single shape that it can bend
at the elbow. Similarly, there is a need for \emph{break-points} at the points of occlusions which cannot be determined
{\em a priori}.  Thus, we 
ensure that at least one breakpoint exists within a certain range of the contour which ensures that
the set of possible GSs has enough number of possibilities that can be used to match with their counterparts
in the other shapes.  Therefore, we also add extra points known as 'max-size points' to our \emph{break-points} to 
handle the case of insufficient number of \emph{break-points}. 
These are added in such a way that the geodesic distance $d_{k}$ between consecutive \emph{break-points}  is 
maintained.  It is taken to be a fraction of the total number of points on the contour (value of $d_{k}$ is 0.1 used in our experiments).  
The portion between consecutive \emph{break-points} is defined as a \emph{segment}.  
The process of creating possible GSs using these \emph{break-points} is described next.

\subsection{Computing Possible Groups-of-Segments (GSs)}
\label{Computing_possible_parts}
We determine possible GSs by considering portions of the contour between any two \emph{break-points}.
A portion $p_{i,j}$ between any two \emph{break-points} 
$i$ and $j$ is taken to be a possible GS if it satisfies a certain complexity range: $C_{MIN}$ $\le$ $C(p_{i,j})$ $\le$ $C_{MAX}$.
The \emph{Complexity} $C$ is defined as the sum of the angles between consecutive segments constituting the GS.
\begin{equation}
{C(p_{i,j})} = \sum_{l = i }^{j-1}(180 - A(Seg_{l},Seg_{l+1})) \label{Complexity}
\end{equation}
\begin{figure*}
  \centering
  \includegraphics[scale=0.4]{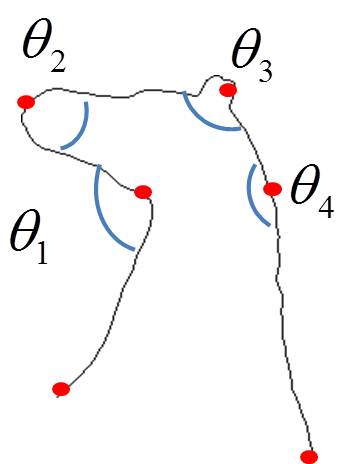}	
  \caption {Angles between adjacent segments for an extracted GS.  The complexity is the sum of such angles: $\theta_{1}$ + $\theta_{2}$ + $\theta_{3}$ + $\theta_{4}$.}  
\label{Complexity_Measure}
\end{figure*}

\begin{figure}
  \centering
  \includegraphics[scale=0.3]{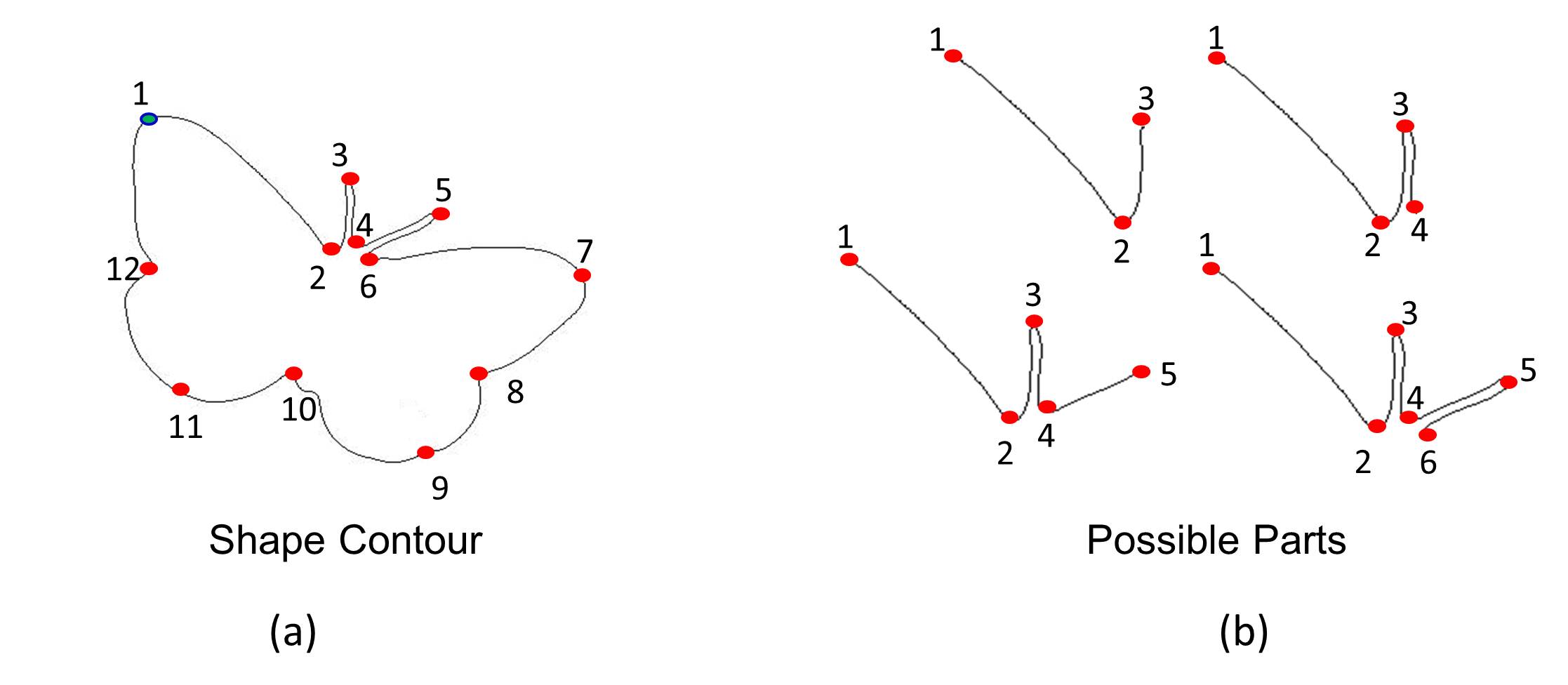}	
  \caption { (a) Possible \emph{break-points} on a butterfly contour. (b) Some possible GSs obtained for (a).} 
\label{Dynamic_merge}
\end{figure}

Eq. \ref{Complexity} measures the Complexity of a GS $gs_{i,j}$.  
$A(Seg_{i},Seg_{i+1})$ gives the angle between the segments, $Seg_{i}$ and $Seg_{i+1}$, where the angle is calculated
from the lines joining the end points of the segments.  Figure \ref{Complexity_Measure} shows an example of such a computation.  
A less complex GS is too simple to match and can match with anything
whereas considering a highly complex GS leads to rigid matching and a very significant computational expense while matching.  
Thus, the range of $C_{MIN}$ and $C_{MAX}$ helps in choosing a subset of all GS possibilities 
that is computationally efficient and is sufficient for most cases.  
Values of $C_{MIN}$ = 40 and $C_{MAX}$ = 600 are used for the experiments in this paper.
Using the above complexity limit, some GS examples thus extracted: $p_{1,3}$, $p_{1,4}$, $p_{1,5}$ and $p_{1,6}$ 
 for a butterfly shape in Figure \ref{Dynamic_merge}(a) are shown in Figure \ref{Dynamic_merge}(b).  
 
\subsection{Affine Shape Normalization of each GS}
\begin{figure}
  \centering
  \includegraphics[scale=0.45]{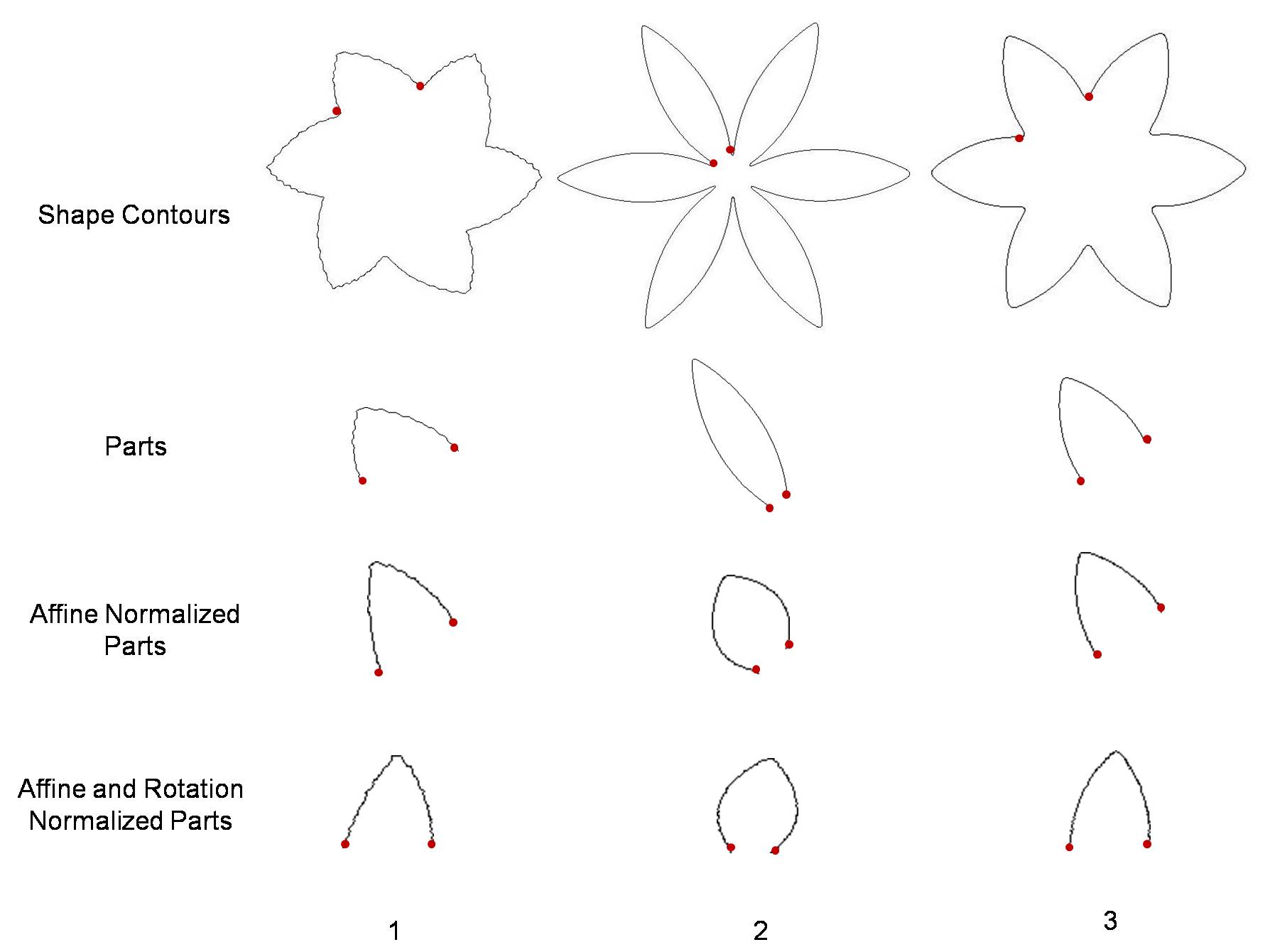}	
  \caption {Affine normalization of GSs of some example shapes for a device-1 category of MPEG-7 dataset \cite{LateckiCVPR200}.}  
\label{Part-Wise-affine-transformations}
\end{figure}

As object and their GSs may appear different in different images due to viewpoint changes or intra-class variations, 
we perform an affine normalization of each GS.  Figure \ref{Part-Wise-affine-transformations} shows an example of 
shape contours that may look globally different due to intra-class variations, but their
affine normalized GSs look quite similar.   Non-rigid deformations that may still exist within the 
 affine normalized GSs are handled by contour matching techniques such as the Fast Directional Chamfer Matching as detailed 
in the next section.

An affine normalization is done using the second order moment matrix $M$ of the contour points: 
\begin{align*}
  \mathrm{M} &= {\begin{bmatrix}\sum \left(x -\bar{x}\right)^2 & \sum \left(x -\bar{x}\right)\left(y-\bar{y}\right) \\
			      \sum \left(y -\bar{y}\right)\left(x -\bar{x}\right) & \sum \left(y - \bar{y}\right)^2
               \end{bmatrix}} = \sum \mathbf{x}_\mu \mathbf{x}_\mu^T  
\end{align*}
By making both the eigenvalues equal, one can estimate the affine-normalization matrix in a manner 
similar to Cohignac et al. \cite{CohignacICPR94} using : %Shen and Ip \cite{DinggangIPPAMI97} using:
\begin{center}
  $\boxed{ \mathbf{A} = \mathbf{M}^{-1/2} }$
\end{center}
Such a normalization corrects the affine transformation, but only up to a rotation as it can also be shown that an 
arbitrary rotation applied to the contour points does not affect the eigenvalues of the Moment Matrix.  
Hence, rotation normalization is applied to make it rotation invariant, using a technique based on whether 
the GS contour is open or closed.  For an open GS contour, 
rotation transformation is estimated by aligning the \emph{start} and the \emph{end} points 
to some fixed points on the horizontal axis.  This handles the scale variation as well. On the other hand, for a closed contour, the rotation 
 is estimated by aligning the \emph{start point} of the contour and the \emph{centroid} of the contour to some fixed points on the horizontal axis 
 as the centroid remains the same under an affine transformation. 

\begin{figure*}
  \centering  
\begin{tabular}{c}
  \includegraphics[scale=0.35]{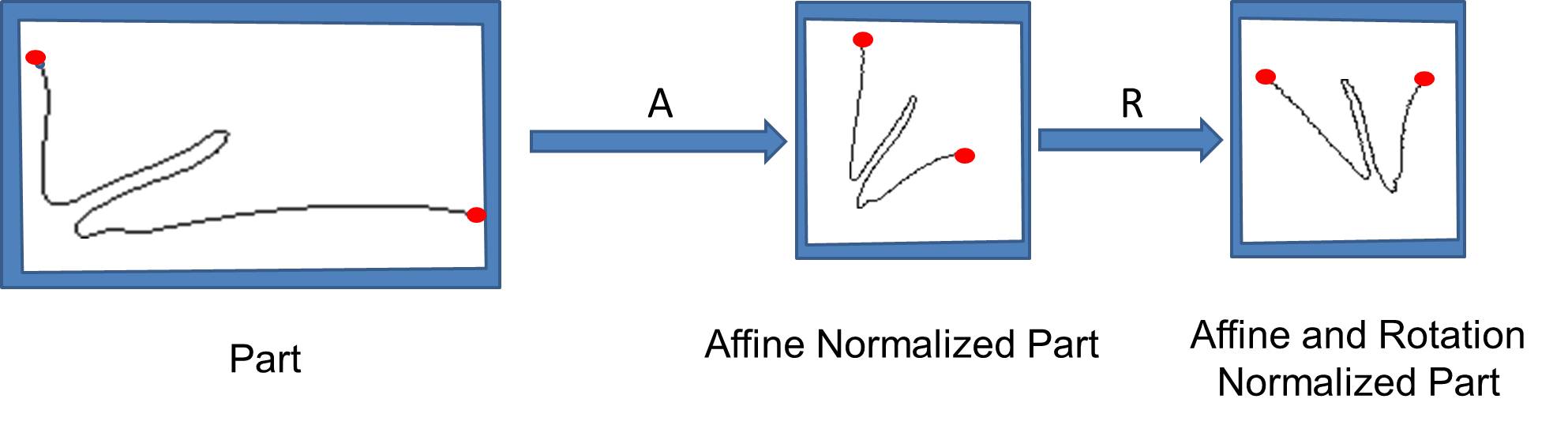}	  \\
  (a)\\
  \includegraphics[scale=0.35]{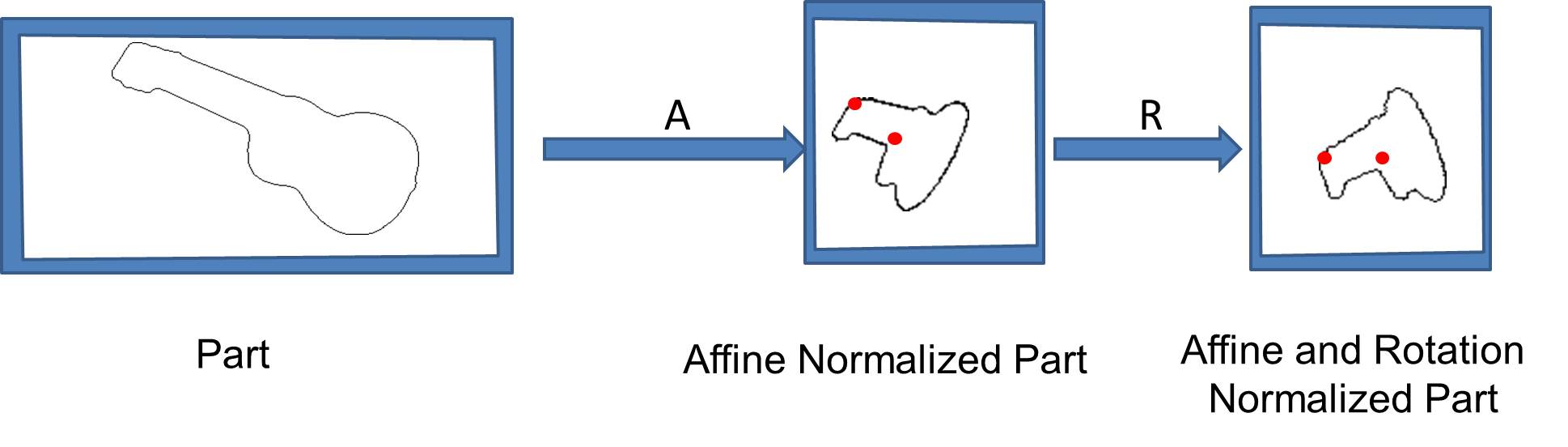}	  \\
(b)
\end{tabular}
\caption{Affine Shape Normalization for (a) an open and (b) a closed GS.}
\label{Affine_shape_Normalization}
\end{figure*}

Such normalization yields locally affine and rotation normalized GSs. Figures \ref{Affine_shape_Normalization}(a) and 
 \ref{Affine_shape_Normalization}(b) show some examples of affine shape normalization for open and closed GS 
contours respectively.  

Given such possible GSs for any two shapes, the cost function for matching them is explained next.

%%%%%%%%%%%%%%%%%%%%%%%%%%%%%%%%%%%%%%%%%%%%%%%%%%%%%%%%%%%%%%%%%%%%%%%%%%%%%%%%%%%%%%%%%%%%%%%%%%%%%%%%%%%%%
%----------------------------------SECTION  3--Shape Similarity---------------------------------------------%
%%%%%%%%%%%%%%%%%%%%%%%%%%%%%%%%%%%%%%%%%%%%%%%%%%%%%%%%%%%%%%%%%%%%%%%%%%%%%%%%%%%%%%%%%%%%%%%%%%%%%%%%%%%%%

\section{Shape Similarity}
Given possible GSs in Images $I_{1}$ and $I_{2}$, 
we define a match between the shapes as a set of GS correspondences across such possible GS sets
that satisfy certain constraints.  Specifically, there must be \emph{non-overlap}, i.e. two matched GSs in an image 
should not intersect with each other.  Next, the order between GSs must be preserved, i.e. if one GS is before 
another GS in an image, such an order must be preserved in the other image.

For a given GS match list (ML) that satisfies these constrains, we define the cost function that evaluates the 
goodness of a match.  How to optimize such a cost function to determine the best
matching will be discussed in later section (Section \ref{Dynamic_Programming_based_matching}).

Let the GS match list $ML$ contain GSs $gs^{1}_{i,j}$ and $gs^{2}_{l,m}$ of Images $I_{1}$ and $I_{2}$ 
respectively and let the entry in $ML$ before $gs^{1}_{i,j}$ and $gs^{2}_{l,m}$ be denoted as $gs^{1}_{prev(i,j)}$ and $gs^{2}_{prev(l,m)}$,
which we refer to as \emph{previous GSs}.  The similarity score for $ML$ is defined using three components: \emph{Unary Cost} ($C_{uc}$), 
\emph{Binary Cost} ($C_{bc}$) and the  \emph{Skip Cost} ($C_{skip}$).

\begin{equation} 
\label{Similarity_Score}
{C(ML)} = \sum_{pair(gs^{1}_{i,j},gs^{2}_{l,m}) \in ML} ( C_{uc}(gs^{1}_{i,j}, gs^{2}_{l,m}) + C_{ipc}(gs^{1}_{i,j}, gs^{2}_{l,m},gs^{1}_{prev(i,j)},gs^{2}_{prev(l,m)}))+ C_{skip}
\end{equation}

\subsection{Unary cost}
The \emph{Unary cost} ($C_{uc}$) has two components: the \emph{Matching Cost} ($C_{match}$) and 
the \emph{Complexity Cost} ($C_{complex}$).  The \emph{Unary Cost} of each GS evaluates how similar and complex these GSs are, considered individually.  
\begin{equation} 
\label{Part_Matching_Cost}
 C_{uc}(gs^{1}_{i,j},gs^{2}_{l,m}) = C_{match}(gs^{1}_{i,j},gs^{2}_{l,m}) +  C_{complex}(gs^{1}_{i,j},gs^{2}_{l,m})
\end{equation}

The \emph{Matching Cost} ($C_{match}$) between the GSs measures their relative similarity.
In this work, we have used \emph{Fast Directional Chamfer Distance (FDCM)} proposed by Liu et al. \cite{FDCM}
for matching the GS contours which works reasonably well as it captures the edge orientation information better compared to 
the traditional \emph{Chamfer matching} \cite{Borgefor1984} or the \emph{Oriented Chamfer matching} \cite{JamieShottonPAMI2008}. 
Such a matching module can deal with non-rigid deformations or noise present in the affine normalized GSs. 
In our experiments, we have empirically chosen the number of orientations as $20$ for 
calculating the directional distance transform.  
To make the matching computationally efficient, we individually precompute the directional distance transform for each GS 
of a shape contour.  Then, the match of another GS with this GS can be computed in an extremely efficient manner. 

The average FDCM score between affine normalized versions of $gs^{1}_{i,j}$ and $gs^{2}_{l,m}$ is represented as 
 $C_{dc}(gs^{1}_{i,j},gs^{2}_{l,m})$ and the \emph{Matching Cost} $C_{match}$ between them is defined as:
%The \emph{Matching Cost} between parts $gs^{1}_{i,j}$ and $gs^{2}_{l,m}$ is defined as:

\begin{equation}
C_{match}(gs^{1}_{i,j},gs^{2}_{l,m}) = w_{p_{(i,j),(l,m)}}\cdot C_{dc}(gs^{1}_{i,j},gs^{2}_{l,m}) \label{Unary_Cost} \\ %\cdot (w^{1}_{i,j} + w^{2}_{l,m}) 
\end{equation}
where
\begin{eqnarray}
&& w_{p_{(i,j),(l,m)}}=w^{1}_{i,j} + w^{2}_{l,m} \label{combine_weight} \ \  \mbox{where}\\
&& w^{k}_{i,j} = \frac {N(gs^{k}_{i,j})} {N^{k}} \label{Weight_part}
\end{eqnarray}
where $w^{1}_{i,j}$ and $w^{2}_{l,m}$ are the weights 
for the matched GSs $gs^{1}_{i,j}$ and $gs^{2}_{l,m}$ in Images $I_{1}$ and $I_{2}$ respectively and 
it can be seen that the weights normalize to $1$: $w^{k}_{n,1} + \sum_{j=1}^{n^{k}-1} w^{k}_{j,j+1} = 1$.  $n^{k}$ represents the number of \emph{break-points} in Image $k$,
$N(gs^{k}_{i,j})$ represents the number of contour points in GS ${gs^{k}_{i,j}}$ whereas ${N^{k}}$ represents the 
total number of contour points in Image $k$.
Such a weighting scheme helps in bringing different matches to the same scale regardless of the number of 
contour points in each shape or the number of GSs in a match.

The matching cost alone does not tell the full picture as some GSs are easier to match than other.  Hence, we introduce a \emph{Complexity Cost}, 
where the \emph{complexity} is defined as before.  This helps in choosing the GSs that are relatively more complex and 
discriminative in terms of their shape structures.  The \emph{Complexity Cost} between GSs $gs^{1}_{i,j}$ and $gs^{2}_{l,m}$ 
is defined as:

\begin{equation}
 C_{complex}(gs^{1}_{i,j},gs^{2}_{l,m}) = w_{gs_{(i,j),(l,m)}}\cdot (\frac {\alpha_{c}}{C(gs^{1}_{i,j})} + \frac{\alpha_{c}}{C(gs^{2}_{l,m})}) \label{Complexity_Measure_eq} \\ 
\end{equation}
where $w_{gs_{(i,j),(l,m)}}$ is the weight of the GSs obtained from Eq. \ref{combine_weight}.  
$C(gs^{k}_{i,j})$ represents the Complexity measure for a GS $gs^{k}_{i,j}$ (the same as in Section \ref{Computing_possible_parts}, Eq. \ref{Complexity}).  
Such a definition defines the complexity of matching a GS much better than simply counting
 the number of segments in it.  The value of $\alpha_{c}$ that controls the weight given to the 
\emph{Complexity Cost} is empirically set to $300$ in our experiments.  

\begin{figure}
\centering
\begin{tabular}{cc}
\includegraphics[height=3cm,width=6.5cm]{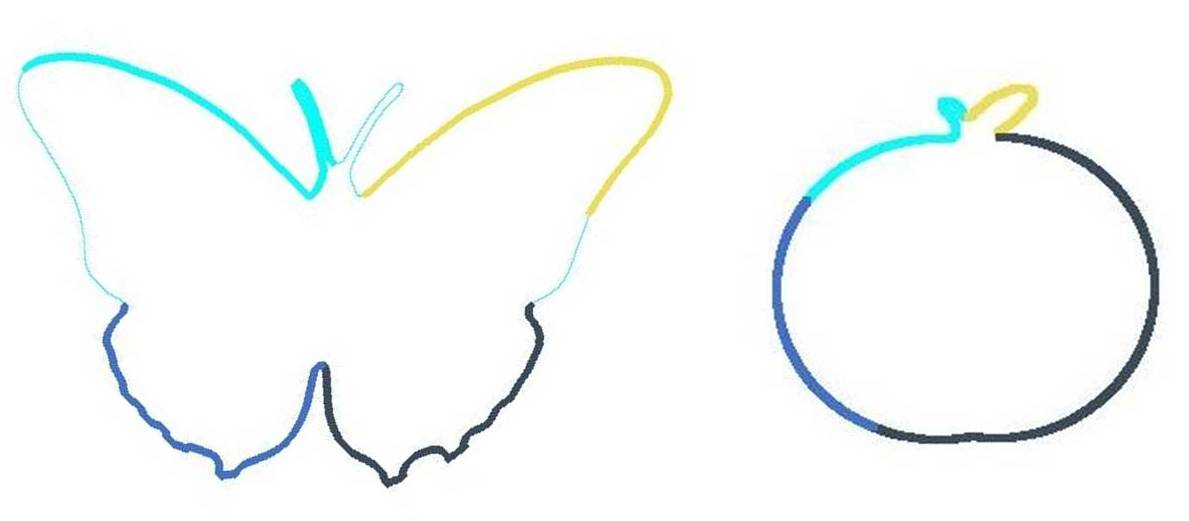}&
\includegraphics[height=3cm,width=6.5cm]{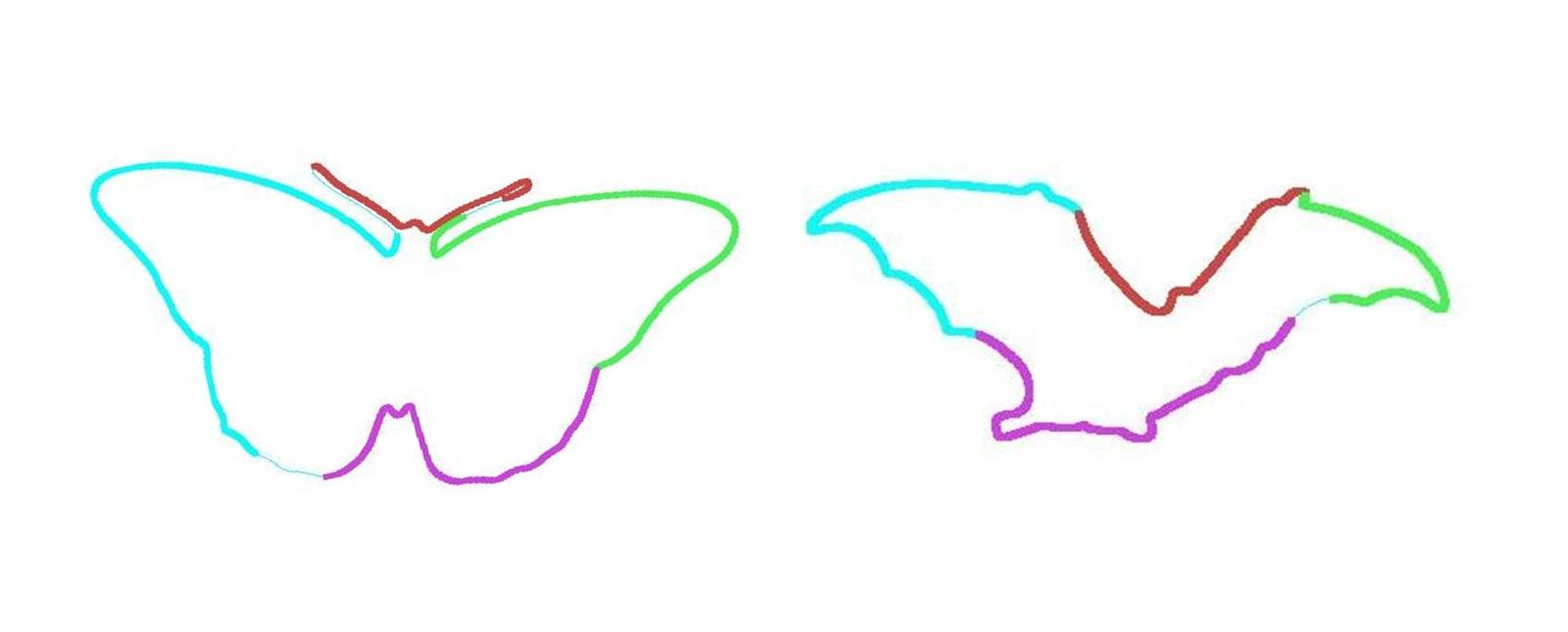} \\
(a) & (b)\\
\end{tabular}
\caption {Best viewed in color.  Some example shapes look globally quite different but their affine-corrected GSs are similar.   
Note that non-matching portions are skipped, which are shown with a light blue color.}  
\label{Part_wise_similar_globally_different}
\end{figure}

The \emph{Unary Cost} ($C_{uc}$) as defined in Eq. \ref{Part_Matching_Cost} helps in choosing sufficiently 
similar, complex GSs between the two images.  Generally, the \emph{Unary Cost} is sufficient to 
distinguish between two shapes under articulations and viewpoint changes.  
However, there are cases where the individual GSs are similar even though the shapes themselves are 
globally different.  Some cases are shown in Figure \ref{Part_wise_similar_globally_different}.  
It may therefore be useful to consider some constraints between adjacent GSs as too 
much size variation or too much articulation at the joint points can lead to a significant distortion in the shape.  
This is considered next, although the weight for these factors is taken to be much less than the unary costs
in our standard implementation, although this can be easily varied as per the requirements of a given application.

\subsection{Binary Cost}
 \emph{Binary Costs} account for Angle and Scale Inconsistencies 
between adjacent GSs.  Enforcing the preservation of these consistencies helps in limiting
the possible articulations at the joint points of these GSs that can sometimes change the shape perspective very 
drastically. 
The \emph{Binary Cost} does not give full global shape matching perspective like some other methods which 
enforce global constraints, but does so in a soft way
giving a shape a second-level perspective beyond individual GS level which helps in improving the results in 
most cases. The relative weights of these costs can be varied as per the shape variations present in a given
application.  For more deformable shapes, a much lower cost for binary costs compared to the unary
costs should be applied to ensure that flexibility while for more rigid shapes, unary costs should be given a higher weight.

We define the \emph{Binary Cost} as:
\begin{equation}
\begin{split}
 \label{Inter-part constraints}
 C_{bc}(gs^{1}_{i,j}, gs^{2}_{l,m},gs^{1}_{prev(i,j)},gs^{2}_{prev(l,m)}) = C_{s}(gs^{1}_{i,j},gs^{2}_{l,m},gs^{1}_{prev(i,j)},gs^{2}_{prev(l,m)}) + \\C_{a}(gs^{1}_{i,j},gs^{2}_{l,m},gs^{1}_{prev(i,j)},gs^{2}_{prev(l,m)}) 
\end{split}
\end{equation}

\emph{Scale Inconsistency} $C_{s}$ at the joint between adjacent GSs is evaluated by measuring the changes in the scale ratios 
of consecutive GSs.  The scale ratio of consecutive GSs is a more appropriate criterion for representing shapes
than using the scale of the whole shape for normalization since the latter may be unreliable especially in the presence 
of occlusions or noise.  The \emph{Scale Inconsistency} for two consecutive pairs of GSs is defined as:
\begin{equation} 
\label{Scale_constraint}
C_{s}(gs^{1}_{i,j},gs^{2}_{l,m},gs^{1}_{prev(i,j)},gs^{2}_{prev(l,m)})   =  \alpha_{s} \cdot w_{ip_{(i,j),(l,m)}}\cdot (1 - e^{-\beta_s \cdot {\bigtriangleup s_{(i,j),(l,m)}}}) 
\end{equation}
where
\begin{eqnarray}
&& w_{ip_{(i,j),(l,m)}} =  w^{1}_{prev(i,j)} + w^{1}_{i,j}+ w^{2}_{prev(l,m)}+ w^{2}_{l,m} \\
&&{\bigtriangleup s_{(i,j),(l,m)}} = abs(\frac {N(gs^{1}_{prev(i,j)})} {N(gs^{1}_{i,j})}  - \frac {N(gs^{2}_{prev(l,m)})} {N(gs^{2}_{l,m})}) \label{scale_ratio_change}
\end{eqnarray}
Here, ${\bigtriangleup s_{(i,j),(l,m)}}$ 
represents the change of relative scale between consecutive GSs: $gs^{1}_{i,j}$ and $gs^{1}_{prev(i,j)}$ in $I_{1}$ 
and their corresponding GSs in $I_2$.  Thus, from Eq. \ref{scale_ratio_change}, we see that large changes 
in the relative scale of adjacent corresponding GS pairs incur a large penalty.  Weights are as defined in Eq. \ref{Weight_part}.

\begin{figure*}
  \centering  
\begin{tabular}{cc}
  \includegraphics[scale=0.28]{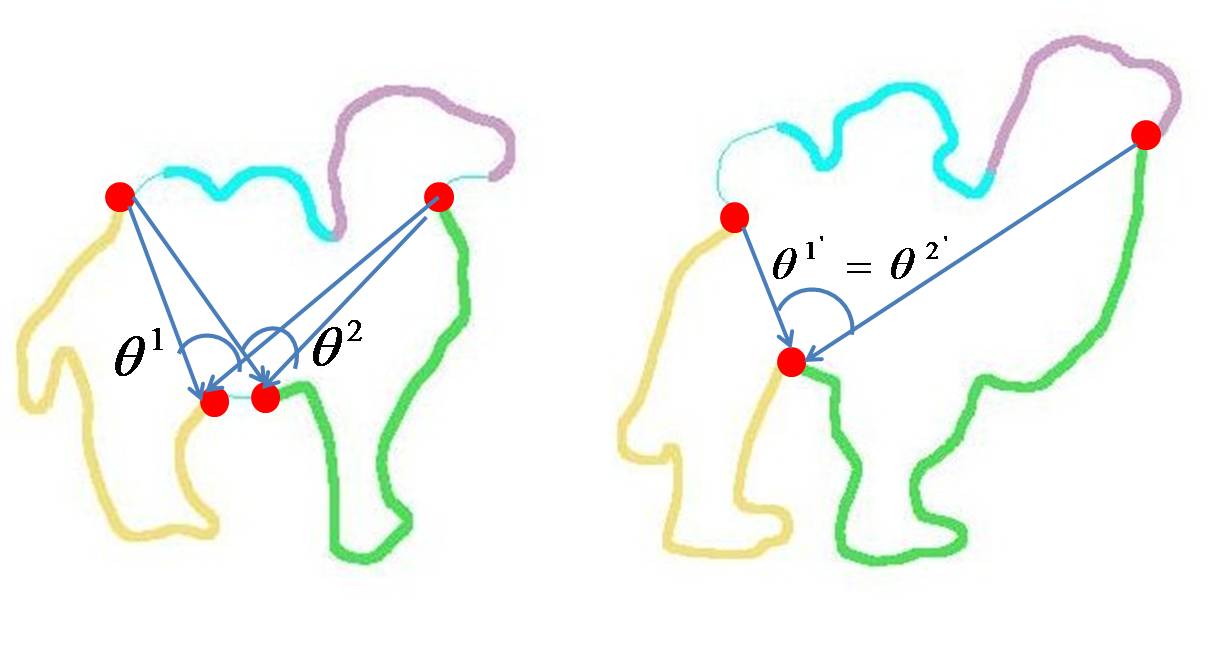}	  &  
  \includegraphics[scale=0.28]{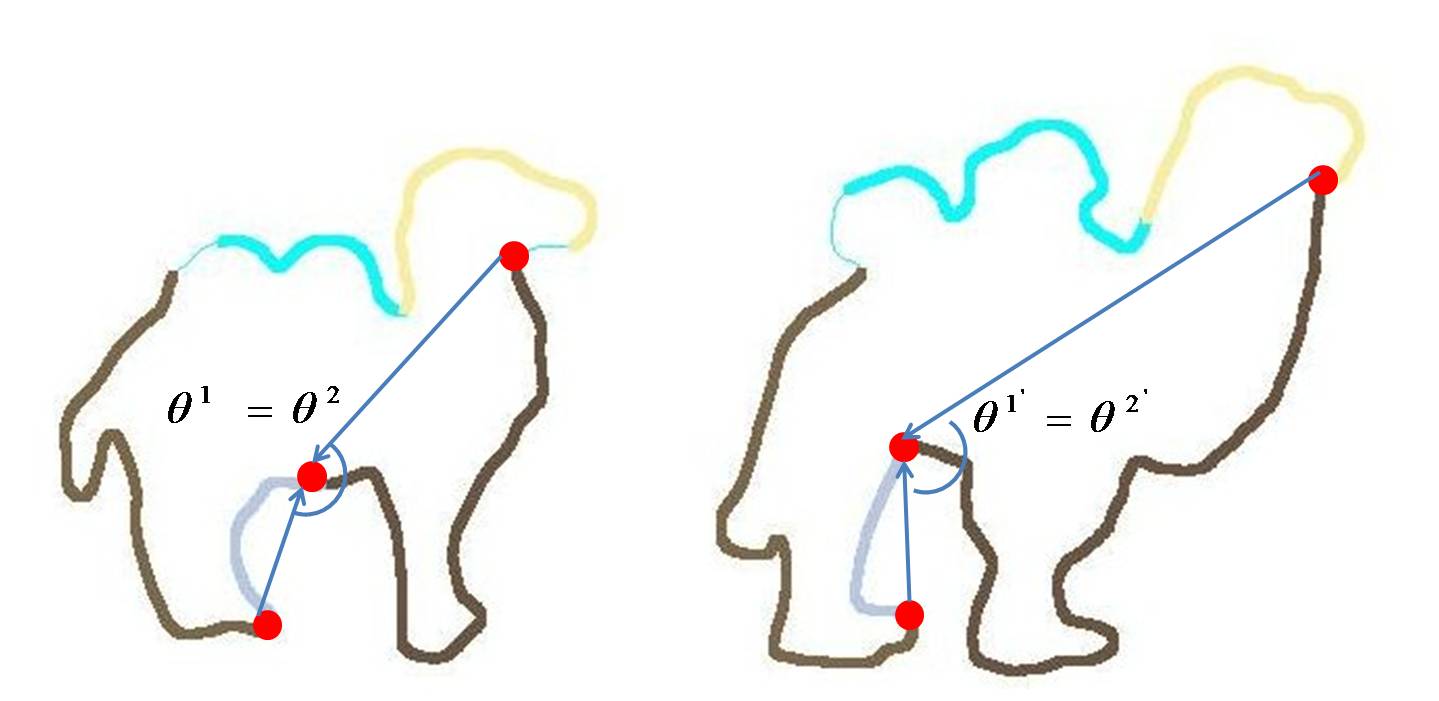}	  \\ 
 (a) & (b)
\end{tabular}
  \caption {Best viewed in color.  GS correspondences (a) with scale and angular constraints, (b) without scale and angular constraints.  
   Note that non-matching portions are skipped, which are shown with a light blue color.}
\label{Part_Correspondence}
\end{figure*}

Similarly, the \emph{Angular Inconsistency} is defined as:
\begin{equation}
\label{Angle_constraint}
C_{a}(gs^{1}_{i,j},gs^{2}_{l,m},gs^{1}_{prev(i,j)},gs^{2}_{prev(l,m)})   =  \alpha_{a} \cdot w_{ip_{(i,j),(l,m)}}\cdot (1 - e^{-\beta_a \cdot {\Delta \theta_{{i,j},{l,m}}}})
\end{equation}
where, $\Delta \theta_{{i,j},{l,m}}$ =  max($|\theta^{1}_{prev(i,j),(i,j)} - \theta^{1'}_{prev(l,m),(l,m)}|$, $|\theta^{2}_{prev(i,j),(i,j)} - \theta^{2'}_{prev(l,m),(l,m)}|$)
represents the change in the relative angle between consecutive GSs $gs^{1}_{i,j}$ and $gs^{1}_{prev(i,j)}$ of Image $I_{1}$ 
with respect to the corresponding GSs in Image $I_{2}$.  The method of measuring $\theta^{1}_{prev(i,j),(i,j)}$, $\theta^{2}_{prev(i,j),(i,j)}$ 
in Image $I_{1}$ and their corresponding angles $\theta^{1^{'}}_{prev(l,m),(l,m)}$ and $\theta^{2^{'}}_{prev(l,m),(l,m)}$ in Image $I_{2}$ is shown 
in Figure \ref{Part_Correspondence}.   Constant values of $\alpha_{a}$ = 200, $\alpha_{s}$ =200, $\beta_a$ = 0.09 and $\beta_s$ = 1.5
are used in the experiments.

\subsection{Skip Cost}
\begin{figure}
  \centering
  \includegraphics[scale=0.5]{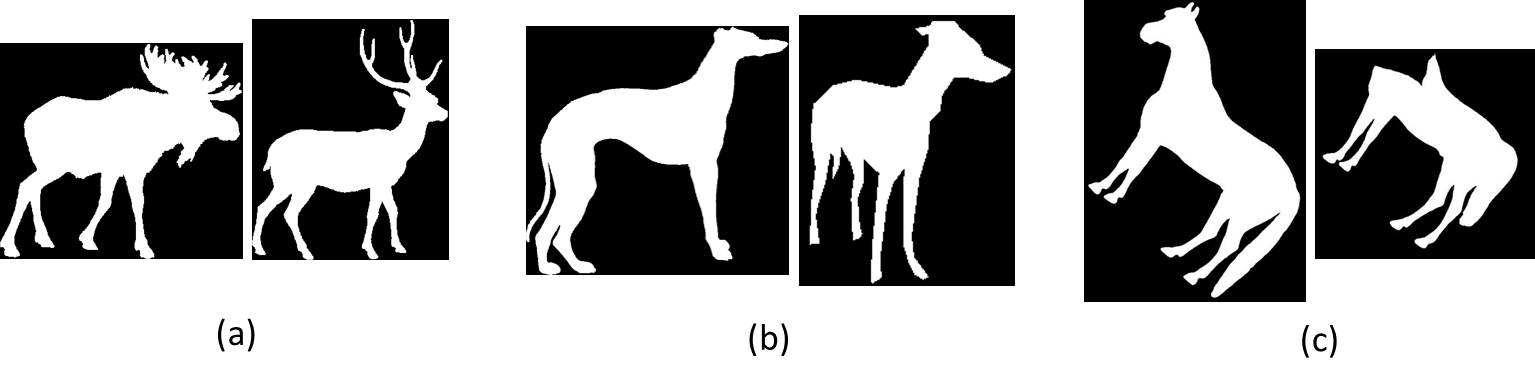}	
  \caption {Portions of shapes can be very different in two matching shapes due to (a) intra-class variation, (b) viewpoint change and (c) occlusions and errors in shape extraction.}  
\label{Shape_variation}
\end{figure}

There are scenarios where some portions of the contour can be missing in either one
or both of the images.  Such scenarios arise due to self-occlusions and/or viewpoint
changes.  Furthermore, for certain categories, some portions of the contour look totally different due to intra-class shape variations.  
Also, shape contours can appear as combinations of multiple other shapes when the
input contours are obtained using segmentation or Background Subtraction techniques.  
These cases are demonstrated in Figure \ref{Shape_variation}.  
To handle such challenges, \emph{Skip Cost} ($C_{skip}$) that facilitates partial matching of the shapes
is considered.

The \emph{Skip Cost} is a penalty for segments that have been skipped because they
do not have a match.  It is calculated using the skipped contours in both the images as:
\begin{equation}
\label{skip_cost}
 C_{skip} = C_{skip}^{1} + C_{skip}^{2} \\
\end{equation}
where
\begin{equation}
 C_{skip}^{k} = \sum_{Seg_{i} \in SSL^{k}} \beta_{skip} \cdot \omega^{k}_{i}   
\end{equation}
where $SSL^{k}$(\emph{Skipped Segments List}) is the list that contains the segments in Image $k$
which do not have a match and the weight $\omega^{k}_{i}$ is as defined before.  Value of $\beta_{skip}$ = $210$ is used in our experiments.

Given the possible GSs in Images $I_{1}$ and $I_{2}$, the procedure for extracting the 
best matching list between two given shapes in terms of the energy function is described next.  

%%%%%%%%%%%%%%%%%%%%%%%%%%%%%%%%%%%%%%%%%%%%%%%%%%%%%%%%%%%%%%%%%%%%%%%%%%%%%%%%%%%%%%%%%%%%%%%
%    Section 4: Dynamic Programming based Matching--------------------------------------------%
%%%%%%%%%%%%%%%%%%%%%%%%%%%%%%%%%%%%%%%%%%%%%%%%%%%%%%%%%%%%%%%%%%%%%%%%%%%%%%%%%%%%%%%%%%%%%%%
\section{Dynamic Programming-based Matching}
\label{Dynamic_Programming_based_matching}
The problem of contour matching is defined in terms of determining a Match List that 
minimizes the energy function.   The value of the energy function of a particular configuration 
depends both on how well the corresponding GSs match and the binary relationships between the consecutively matched GSs.
We realize that the entries of $ML$ (in Eq. \ref{Similarity_Score}) are circular which makes the problem hard.  
However, if one neglects the binary constraints between the last and the first GS, one can break the cycle 
and solve the problem exactly using \emph{Dynamic Programming}(DP).
This approach is used in our work due to its much greater efficiency.
The problem thus obtained is similar to the \emph{Longest Common Sub sequence} (LCS) \cite{CormenBook2001}
problem, where one needs to match a common subsequence of tokens across two given sequences.  However, 
there are some additional considerations due to the \emph{non-overlap} and the \emph{binary} 
constraints between the matched GSs.  

More precisely, let the sets of possible GSs of Images $I_{1}$ and $I_{2}$ be $\mathcal{GS}_{1}$ and 
$\mathcal{GS}_{2}$ respectively.  Then, we require a one-to-one mapping from $\mathcal{GS}_{1}$ to $\mathcal{GS}_{2}$ in an {\em order-preserving} manner such that the matched GSs should not have any overlapping segments between them.  Using the cost function defined in the 
previous section, we define the best matching between two contours as:

\begin{equation} 
\label{Minimize_Energey}
{ML^{*}} = arg \displaystyle \min_{ML} \Big(\sum_{pair(gs^{1}_{i,j},gs^{2}_{l,m}) \in ML} ( C_{uc}(gs^{1}_{i,j}, gs^{2}_{l,m}) + C_{ipc}(gs^{1}_{i,j}, gs^{2}_{l,m},gs^{1}_{prev(i,j)},gs^{2}_{prev(l,m)}) + C_{skip} \Big)
\end{equation}
wherfe $ML$ is a match list that contains matched GS correspondences as defined in the previous section.  
To solve the problem exactly for the cost function defined in Eq. \ref{Minimize_Energey}, one needs to search over 
all possible GS combinations.  The solution space becomes very large if one considers all possible skips between the GSs while 
maintaining order constraints.
For that, one would need to build a 2D matrix $T$ such that each element 
$T(gs^{1}_{i,j},gs^{2}_{l,m})$ represents the minimum total cost of matching the two shapes 
up to the matching GS-pair ($gs^{1}_{i,j}$,$gs^{2}_{l,m}$), to compute which, one would have to search over a large set of combinations of
the previous matching GS pairs $(gs^{1}_{prev(i,j)},gs^{2}_{prev(l,m)})$ due to the possibility of skipped segments.
The running time of such an algorithm would be $M \times N \times S^2$, where $M$ and $N$ are possible number of GSs in 
Images $I_{1}$ and $I_{2}$ respectively and $S$ is the maximum skip allowed.

This is quite expensive and impractical and hence, we consider an approximate solution to the problem 
that is much faster and works reasonably well in practice.  The approximation is done by minimizing the cost function at a 
block level.  A block is considered as a set of possible GS correspondences whose ending \emph{break-points} are the same.  
Only the best match within a block is saved for the next step in the optimization.  
This approximation helps in reducing the number of possible skips from quadratic to linear.
More precisely, we propose a memory-efficient 
solution where we store only an $m \times n$ matrix $T$ such that
each element $T(i,j)$ represents the minimum total cost of matching between Images $I_{1}$ and $I_{2}$ up to the matches of the 
$i^{th}$ and the $j^{th}$ segments of Images $I_{1}$ and $I_{2}$ respectively.  Note that since we now work on segments and 
not GSs, the dimensions $m$ and $n$ of the matrix $T$ are the possible number of \emph{break-points} or segments in 
Images $I_{1}$ and $I_{2}$ and not the number of GSs which is much higher.   

\begin{figure*}
  \centering  
  \includegraphics[scale=0.6]{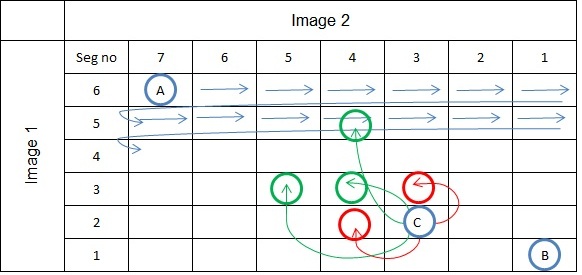}	
  \caption {Best viewed in color.  Skipping is shown in read circle and some of the possible matches are shown in green color. }
\label{DP_memory_organization}
\end{figure*}

The minimum total cost of matching 
up to the entry $(i,j)$ can be calculated using Dynamic Programming using the following recurrence relation:
\begin{equation}
\label{gen_eq}
\begin{split}
T(i,j) =\begin{cases}   
     \infty	\hspace{85 mm}  if\   i = 0   \|  j = 0 \\
    min \Big[
\displaystyle \min_{gs^{1}_{i-t,i} \in P^{1}, gs^{2}_{j-q,j} \in P^{2}}
	\Big[T(i-t,j-q)) + C_{uc}(gs^{1}_{i-t,i},gs^{2}_{j-q,j})  \\
		\hspace{38 mm}  + C_{ipc}(gs^{1}_{i-t,i},gs^{2}_{j-q,j},lastGS(i-t,j-q,1),\\\hspace{38 mm} lastGS(i-t,j-q,2)) \Big], \\
	 \hspace{38 mm} T(i-1,j) + skip^{1}_{i},  \\
	 \hspace{38 mm} T(i,j-1) + skip^{2}_{j} \Big] \hspace{25 mm} otherwise \\	 
\end{cases}
\end{split}
\end{equation}
where ($lastGS(i-t,j-q,1)$,$lastGS(i-t,j-q,2)$) is the best matched GS correspondence, 
ending with segment indexes $i-t$ and $j-q$, where $t$ and $q$ represent the number of segments in GSs $gs^{1}_{i-t,i}$ and $gs^{2}_{j-q,j}$ 
of Images $I_{1}$ and $I_{2}$ respectively.  $skip^{k}_{i}$ represents the skip cost for skipping the $i^{th}$ segment in Image $k$. 
The computation of $T(i,j)$ as described above can be arranged in a sequence as shown in Figure \ref{DP_memory_organization}, such that all the necessary 
terms for the calculation of $T(i,j)$ are already available at the time of its computation and is stored in a table 
along with the least total cost at each stage.  We also store the last best matched GS 
correspondence ($lastGS(i-t,j-q,1)$,$lastGS(i-t,j-q,2)$) for each $T(i,j)$, which not only 
helps in the next stage of the algorithm but also to trace the best GS correspondences in the end, 
as is typically done in Dynamic Programming solutions \cite{CormenBook2001}. 

Let $N$ and $M$ be the number of GSs in the Images $I_{1}$ and $I_{2}$ respectively.
Then, the time complexity of matching two shape contours is $O(N\times M)$ using this algorithm when both the shape contours are 
already aligned.  For handling rotations, one has to consider all possible starting points.  Since
we extract \emph{break-points} such that the number of starting points in the Image $I_{1}$ can be
restricted to only the \emph{break-points},  the algorithm is much more efficient compared to other contour
point-based approaches (\cite{HJD,SC}) which have to search over all the contour points for the starting point correspondence.  
Furthermore, the required memory storage in our approach is $O(m\times n)$, where $m$ and $n$ are the number of segments in 
Images $I_{1}$ and $I_{2}$ respectively, which is much less compared to the number of contour points.  
To give some idea of the actual processing time of our DP-based matching, we ran our code on a 64-bit 2.4Ghz single-core i7 processor 
machine for matching $100$ different preprocessed shape pairs of MPEG-7.  Our DP-based matching took $0.355$ second to compare two shapes where 
average number of extracted possible GSs was $126$.  

\begin{comment}
To give some idea of the actual processing time of the algorithm, we ran our code on a 64-bit 2.4Ghz single-core i7 processor machine for matching $100$ different shape pairs of MPEG-7.
The average number of segments was $22$ with $126$ average number of extracted possible GSs.
Pre-processing(once per shape) took on average $7.674s$ with an unoptimized MATLAB code ($1.09s$  (break-point computation) +
$6.584s$  (Directional Distance Transform computation)) and matching two shapes took $0.355s$  (GS matching took $0.143s$  (C++ Code) \&
DP-based matching took $0.212s$ (C++ Code)).
\end{comment}
In the next section, we compare the performance of our algorithm with other existing approaches on 
some standard datasets.  

\section{Experiments}

We evaluate our algorithm for shape matching for the task of shape retrieval
when the objects are represented only by contours or silhouettes.  This is typically the
output from many automatic segmentation techniques such as Background Subtraction or Image Segmentation.  
First, we show results on the popular MPEG-7 shape dataset \cite{LateckiCVPR200} and compare
against other methods that have been considered in the past.  
Apart from the original dataset, we also show results of matching when the shape extraction has some errors
due to missed portions or merged segmentation.  This is done by simulating such errors on the MPEG-7 dataset.
Then, we evaluate our algorithm against IDSC \cite{HJD} and Gopalan et al. \cite{RPRC} on a dataset provided by Gopalan et al. \cite{RPRC}
created using a real Background Subtraction algorithm on different human and robot poses.  
The results obtained on such a dataset would be indicative of the performance of the algorithms 
in scenarios involving real 3D articulated objects. % where those objects are also moving in a scene.
Finally, we show visual results on the 2D Mythological Creatures dataset provided by Bronstein et al. \cite{ABronsteinIJCV2008} that contains shapes obtained  
by artificially merging contours of different creatures.

\subsection{MPEG-7}

The MPEG-7 is a widely used dataset for evaluation of contour-based shape recognition and retrieval methods.
It contains $1400$ images - $20$ shapes per class from $70$ different classes.  
The dataset is challenging because of the presence of deformations, articulations, GS-wise affine 
changes and missed or altered contour segments in the images. 

\subsubsection{Results on the Basic MPEG-7 Dataset}

\begin{figure}
  \centering 
\begin{tabular}{c} 
  \includegraphics[scale=0.28]{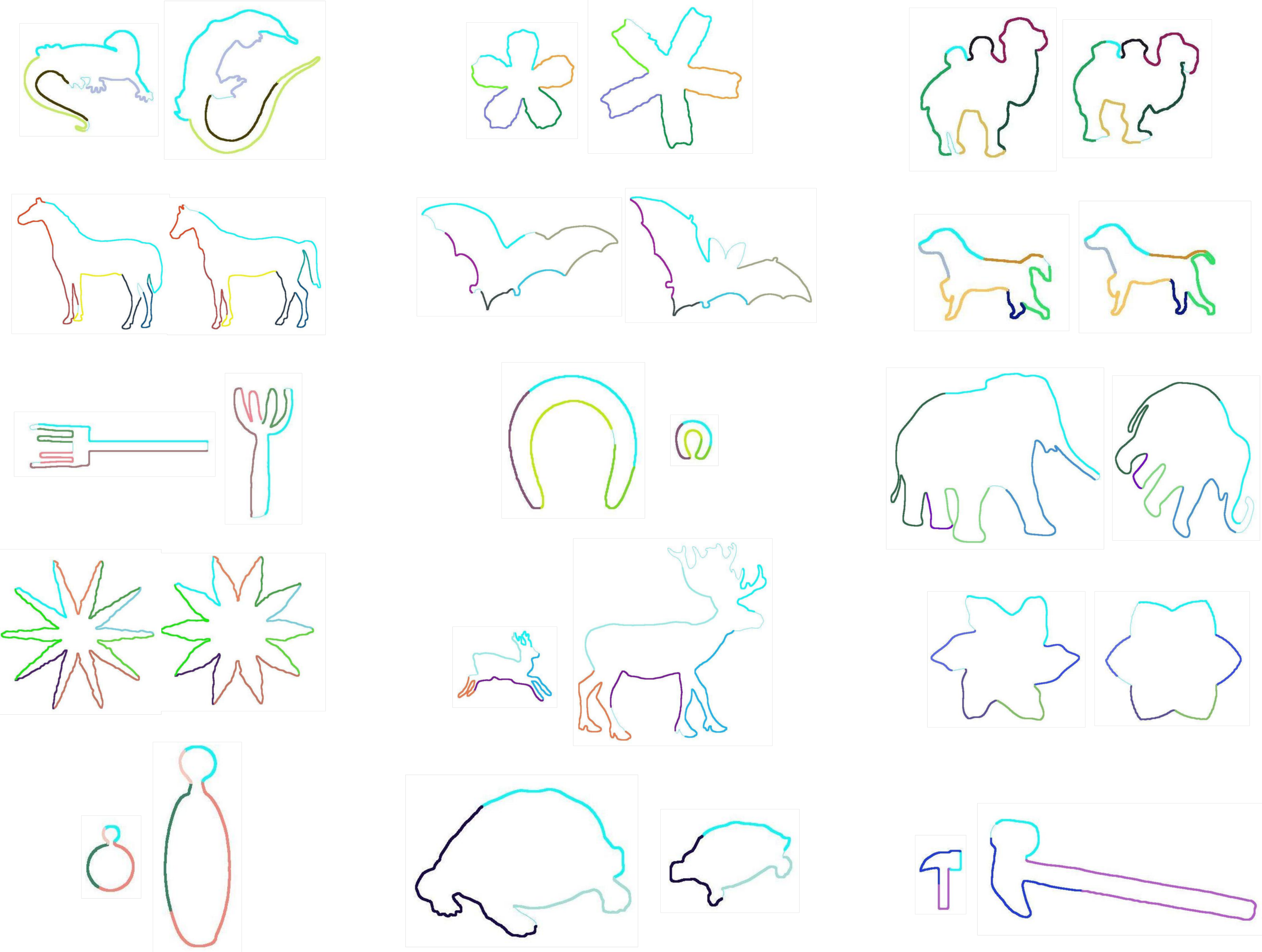}	\\
\includegraphics[scale=0.28]{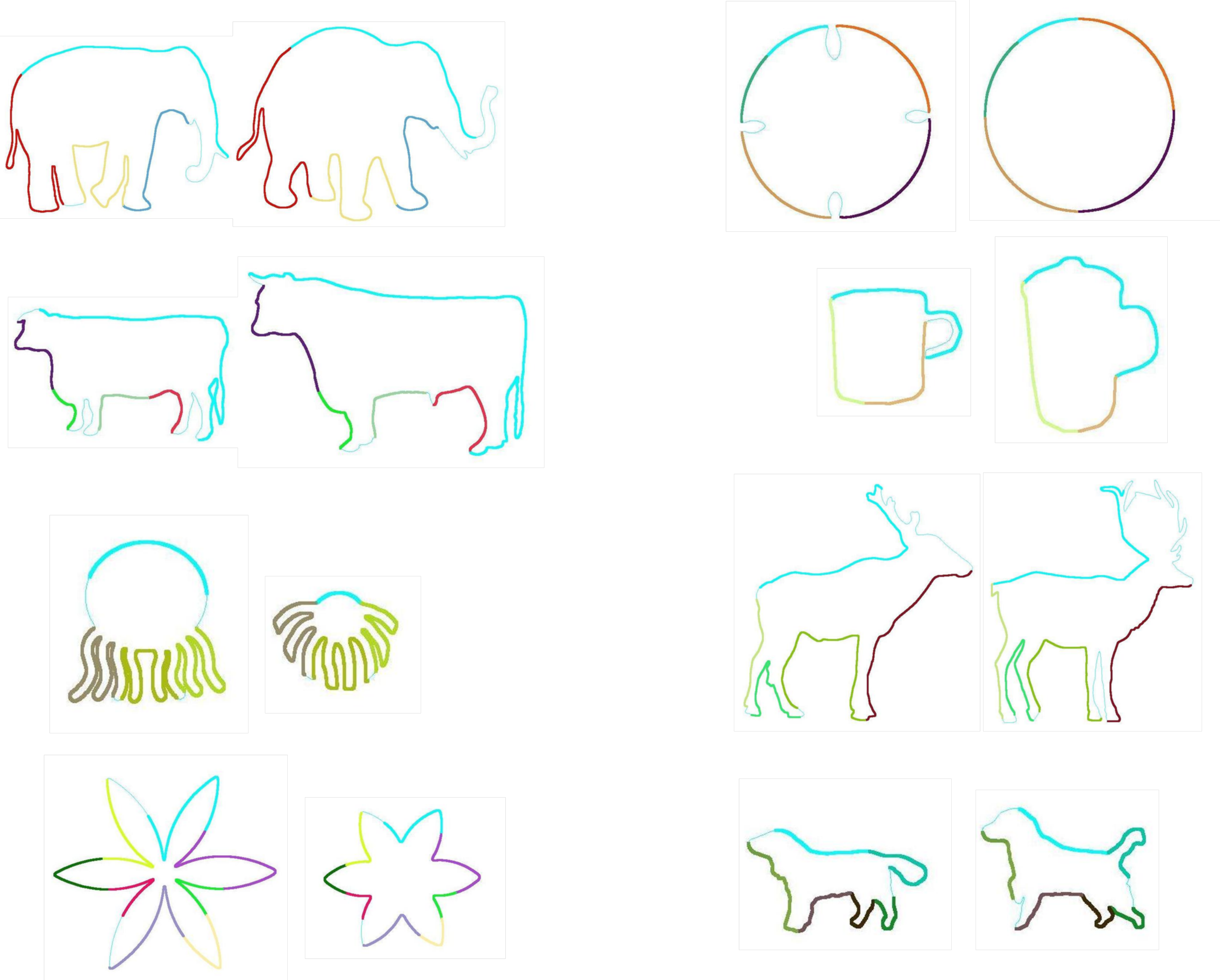}	 \\
\end{tabular}
\caption{Best viewed in color.  Some matchings and best shape decompositions obtained between pairs of shapes.
}
\label{Articulation_Occlusion_MPEG7}
\end{figure}

Figure \ref{Articulation_Occlusion_MPEG7} shows some matching results depicting also the best matching shape-decompositions 
obtained by our algorithm on the MPEG-7 dataset.  
Similar GS color corresponds to a matched GS between the two images.  
Note that non-matching portions are skipped, which are shown with a light blue color.

Table \ref{MPEG7} compares our approach with various existing methods such as \cite{FPS,LingHPAMI2007,ATemlyakovCVPR2010} for the shape retrieval task on the MPEG-7 dataset.  The figures are taken from the respective papers which have reported the Bulleye score for this
dataset.
The proposed method performs reasonably well as compared to many other techniques.    
The methods proposed in \cite{BaiPAMI2010,XKSLL} do not perform individual shape-to-shape matching in isolation, 
but learn the shape variations present in the dataset in order to improve their performance, taking into account not just the similarity
between the shapes of the same category but also the dissimilarity between shapes in different categories in order to train their
matching function.  Thus, it may be claimed that it is unfair to compare our approach with these learning-based methods that depend
heavily on the use of a classified database which may not always be available.  
Furthermore, it is not clear
how the methods would perform on other shapes of a category, such as articulated shapes, if shapes close to those are not present in the
database.  The method proposed by Gopalan et al. \cite{RPRC} performs the best by affine normalizing each of the convex parts before
matching using IDSC.  
\begin{table*}
  \caption{Comparative retrieval results on the entire MPEG-7 dataset \cite{LateckiCVPR200}.}
 \small
\begin{center}
  \begin{tabular}{|l|c|} \hline
\hline\noalign{\smallskip}
            Algorithm & Bullseye Score(in \%)\\ 
	      \hline     
	    Visual Parts \cite{LateckiCVPR200} & 76.45 \\ 
            SC+TPS \cite{SC} & 76.51 \\
	    Curve Edit  \cite{TSebastianPAMI2003} & 78.14 \\
            Generative models \cite{TZYA} & 80.03 \\
	    Curvature scale space \cite{FMokhtarianCSS2003} & 81.12 \\
	    Chance Probability Function \cite{SuperCVPRW2004} & 82.69 \\
	    Fixed Correspondence \cite{BoazIJPRAI2006} & 84.05 \\ 
	    Polygonal Multiresolution \cite{AttallaPR2005} & 84.33 \\
	    Multiscale Representation \cite{AdamekCSVT2004} & 84.93 \\
	    Shape L'\^{A}ne Rouge \cite{peterCVPR2008} & 85.25 \\
	    IDSC + DP \cite{HJD} & 85.40\\
	    Symbolic Representation \cite{DaliriPR2008}  & 85.92 \\ 
	    Hierarchical Procrustes \cite{GMcNeillCVPR2006} & 86.35 \\
	    IDSC + DP +EMD \cite{LingHPAMI2007} & 86.86 \\
	    Triangle area \cite{NAlajlanPAMI2008} & 87.23 \\
            Shape-tree \cite{FPS} & 87.70\\
	    IDSC + AspectNorm. + StrandRemoval \cite{ATemlyakovCVPR2010} & 88.39 \\
	    Contour flexibility \cite{XChunjingPAMI2009} & 89.31 \\
	    Label Propagation \cite{BaiPAMI2010}& 91.61 \\  
	    Locally constrained diffusion \cite{XKSLL} & 93.32 \\
 	    IDSC + Affine Normalization \cite{RPRC}& 93.67 \\ 
	    Ours  &  88.82 \\
\hline
\end{tabular}
\end{center}
\label{MPEG7}
\end{table*}

\begin{figure*}
  \centering  
\begin{tabular}{ccc}
\includegraphics[scale=0.25]{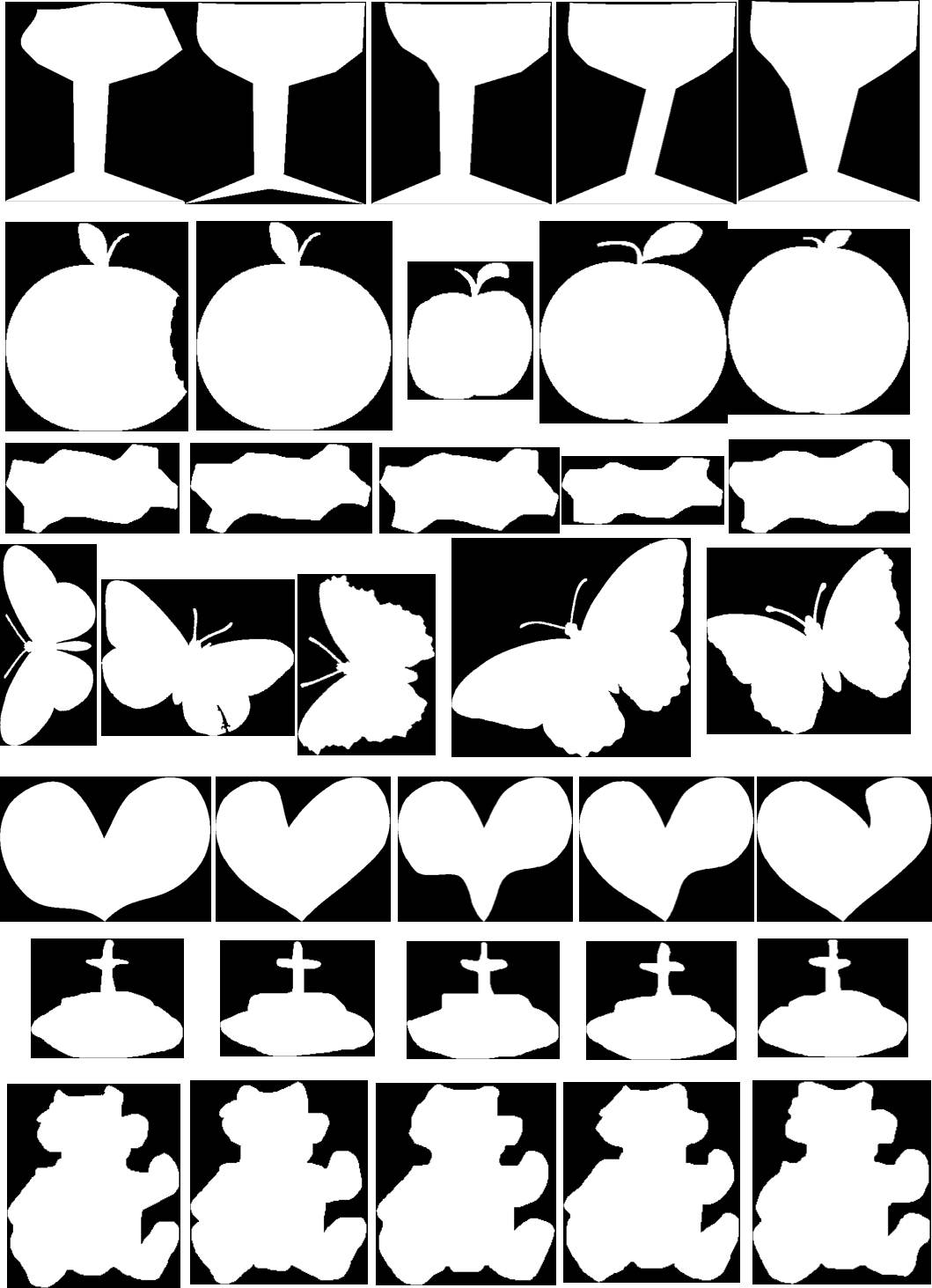}	  &
\includegraphics[scale=0.25]{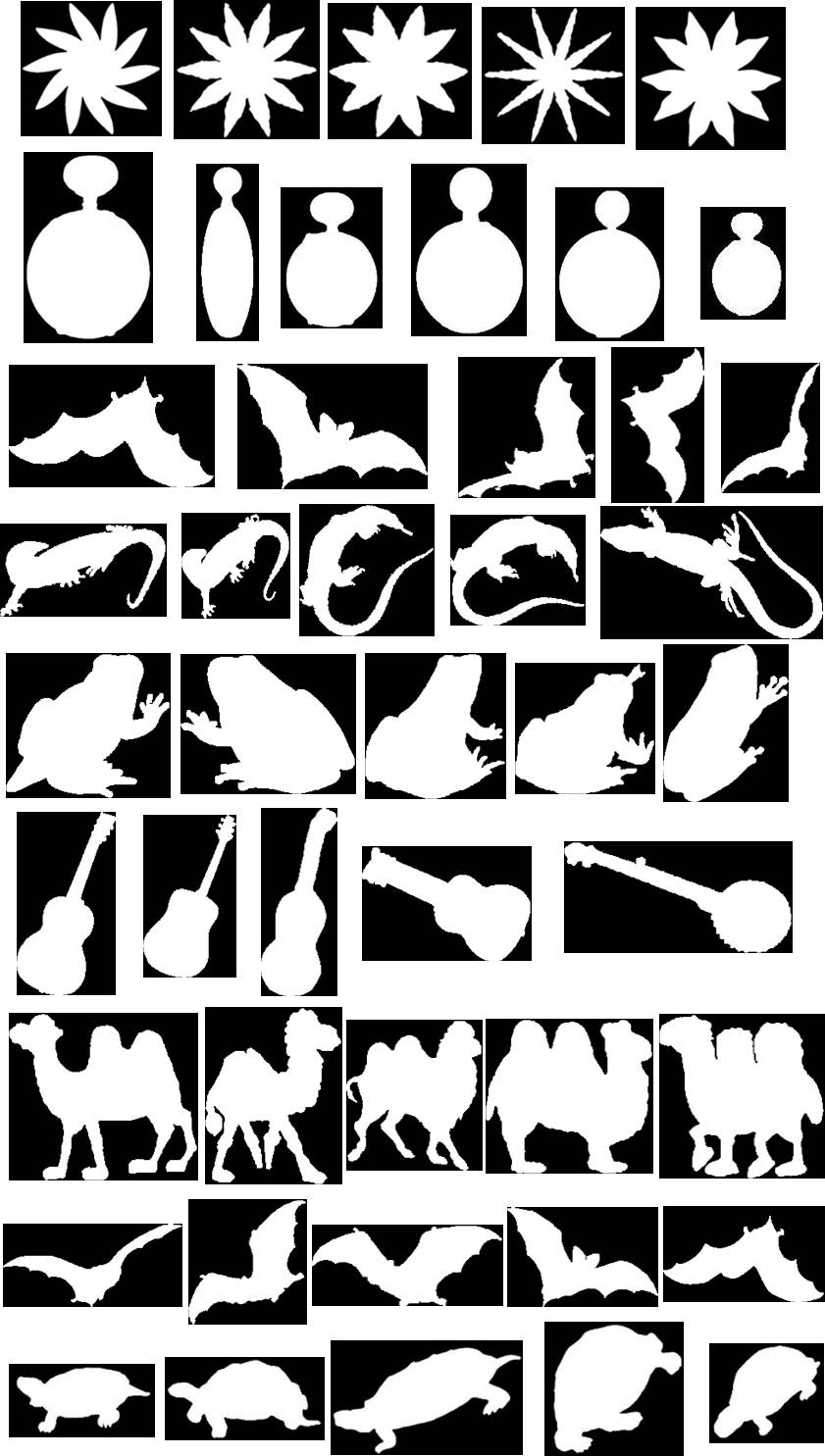}	  &
\includegraphics[scale=0.26]{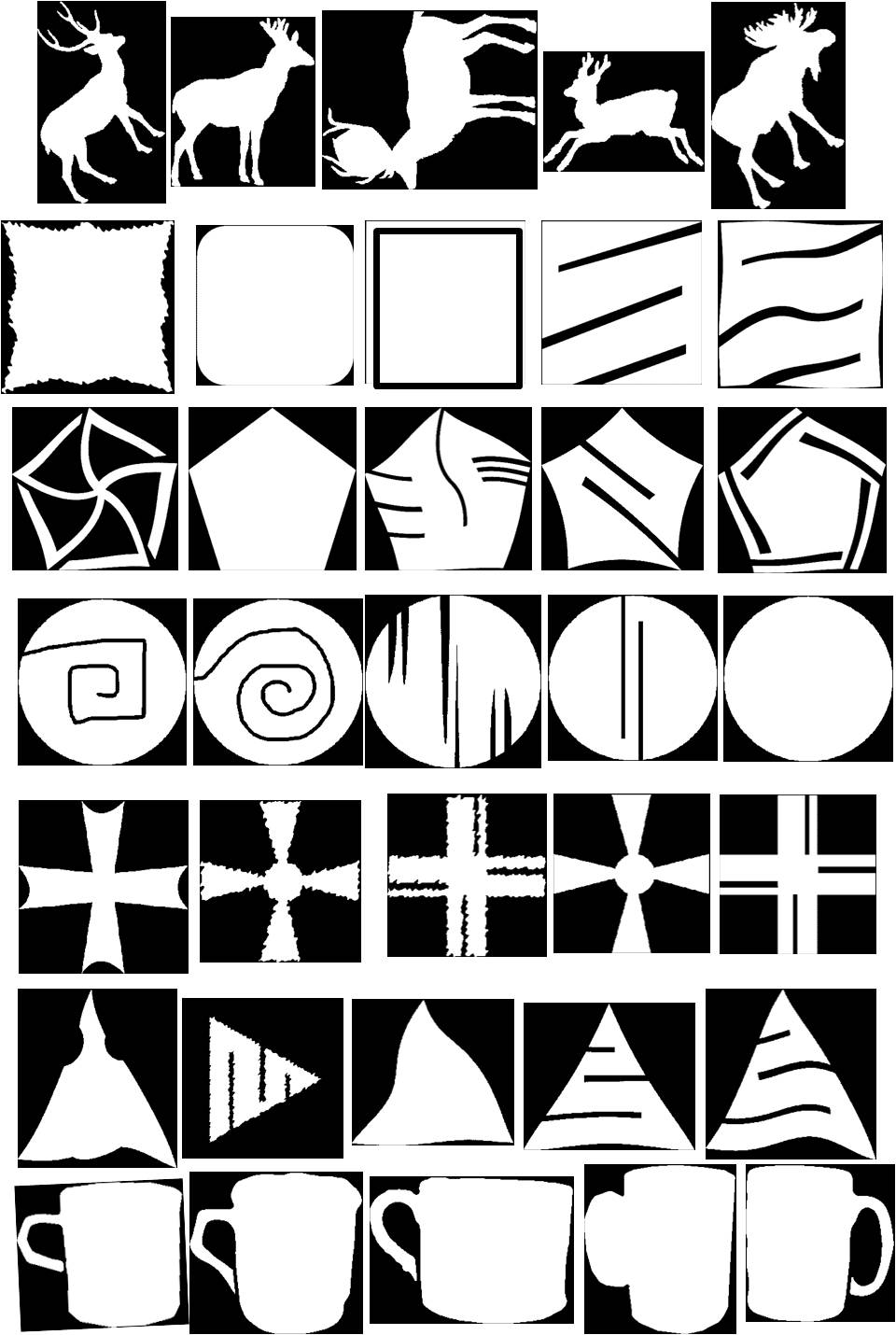} \\
(a) Rigid & (b) Articulated & (c) Missed or altered contour portions. \\
\end{tabular}
\caption{Categorized example shapes from the MPEG-7 dataset.}
\label{Example_Shape}
\end{figure*}

While the results on the entire dataset are interesting, they provide little insight into the strengths and weaknesses of each method and there
is no guarantee that the same ordering of the methods would be obtained on another dataset.
A more refined evaluation may be performed by dividing the dataset into $3$ categories depending on the 
type of the shape variations present: the first containing mostly rigid objects, with possible rotations or scale changes; 
the second containing articulations, deformations and GS-wise affine changes and the last containing missed or altered contour portions.  
Examples of such categories of shapes are shown in Figure \ref{Example_Shape}.   

For these categories, we compare our approach with IDSC and IDSC+Aff \cite{RPRC} for the task of shape retrieval.  
IDSC is used since the code was freely available\footnote{\small{\textit{\url{http://www.dabi.temple.edu/~hbling/code/idsc_distribute.zip}}}} and it ran reasonably fast.  
The implementation of the best reporting one, IDSC+Aff proposed by \cite{RPRC}, is not publicly available.  
Hence, we have re-implemented the paper with the code for shape-decomposition available from the author \footnote{\small{\textit{\url{http://www.umiacs.umd.edu/~raghuram/Segmentation_FULL_NCut.zip}}}} 
and tried to reproduce the results as best as possible, testing for different possible parameters.  
However, we weren't able to reproduce the exact results on the overall dataset as reported by the authors and 
got slightly lower results ($86.9\%$ compared to $93.67\%$).  The category-wise results were not reported in their paper and 
hence the category-wise comparisons as also the comparisons in the next two sections 
(MPEG-7 Merged and Partial Occlusion datasets) are done using our re-implementation.  
However, in spite of such differences in the implementations, the overall pattern of the results should still be the same.  

\begin{figure*}
 \centering  
\begin{tabular}{c}
\includegraphics[scale=0.38]{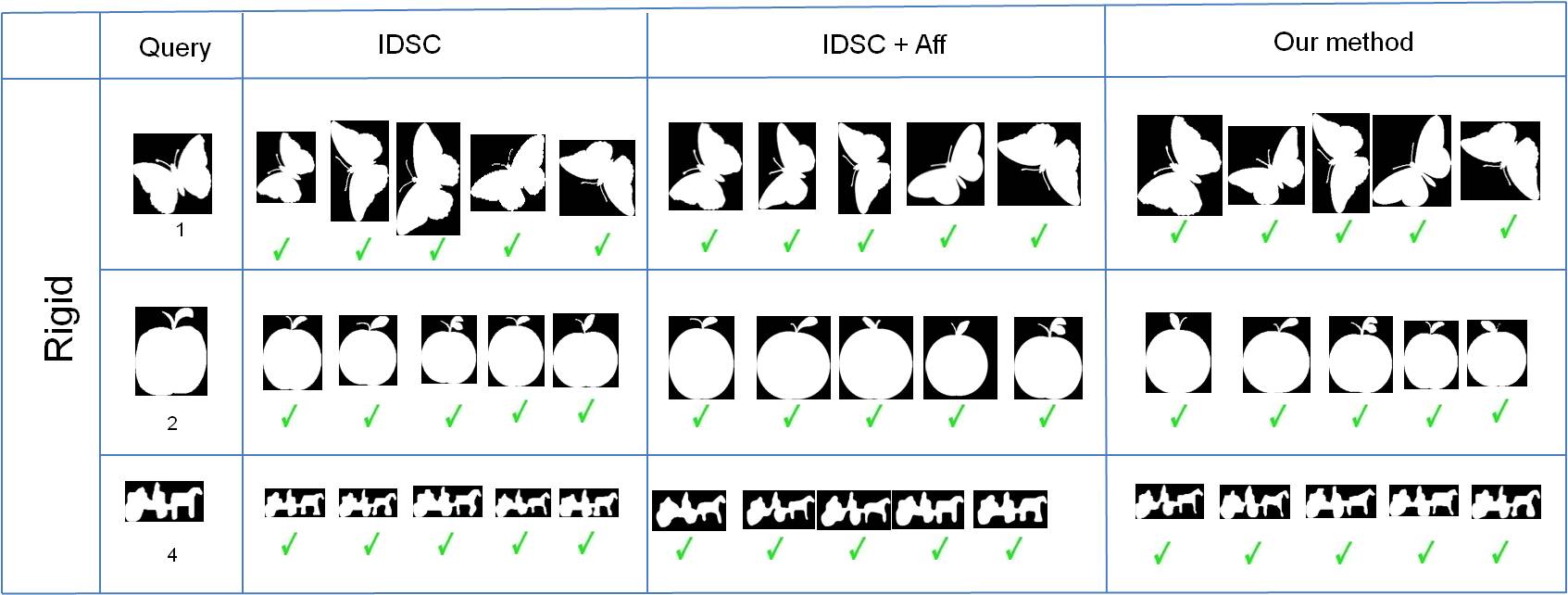}          \\
(a) The 5 most similar shapes retrieved by IDSC, IDSC+Aff and our method.  \\ \\
\includegraphics[scale=0.58]{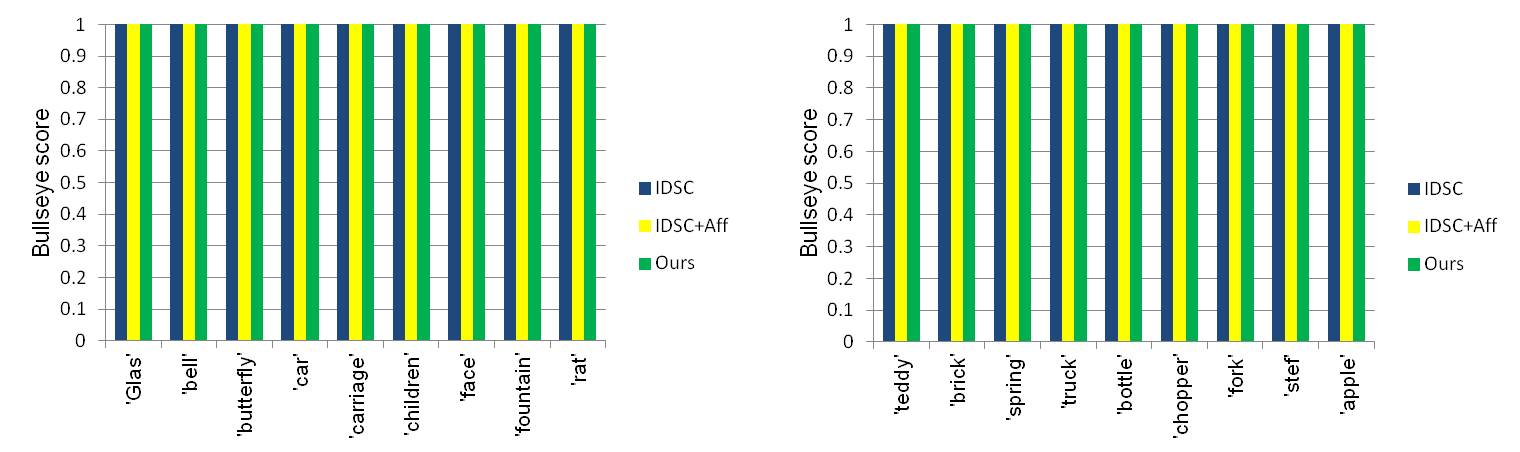}  \\
(b) Shape-wise average Bullseye scores.  \\ Average: IDSC: 100$\%$, IDSC+Aff : 100$\%$, Ours: 100$\%$. \\
\end{tabular}
\caption{Performance of IDSC vs. IDSC+Aff vs. our approach on the rigid shapes of the basic MPEG-7 dataset which contain little intra-class variations. }
\label{Rigid}
\end{figure*}

Figure \ref{Rigid}(a) visually shows the retrieved results while Figure \ref{Rigid}(b)
quantitatively compares the category-wise average Bullseye scores for shapes that are mostly rigid but have some deformations.
The Bullseye score for a query shape is measured by identifying the number of correct retrievals in the top $40$ retrieved shapes.  
As can be seen, all three methods - ours, IDSC and IDSC+Aff - achieve a retrieval rate of $100$ percent on all the rigid query shapes.  
A high performance for this category of shapes for both methods may be attributed to very little intra-class variations.  
Most of the methods, including those that match shapes rigidly such as the Chamfer Distance (\cite{Borgefor1984}) or the 
Hausdroff Distance (\cite{HD}), should do well on this category of shapes.

\begin{figure*}
 \centering  
\begin{tabular}{c}
\includegraphics[scale=0.4]{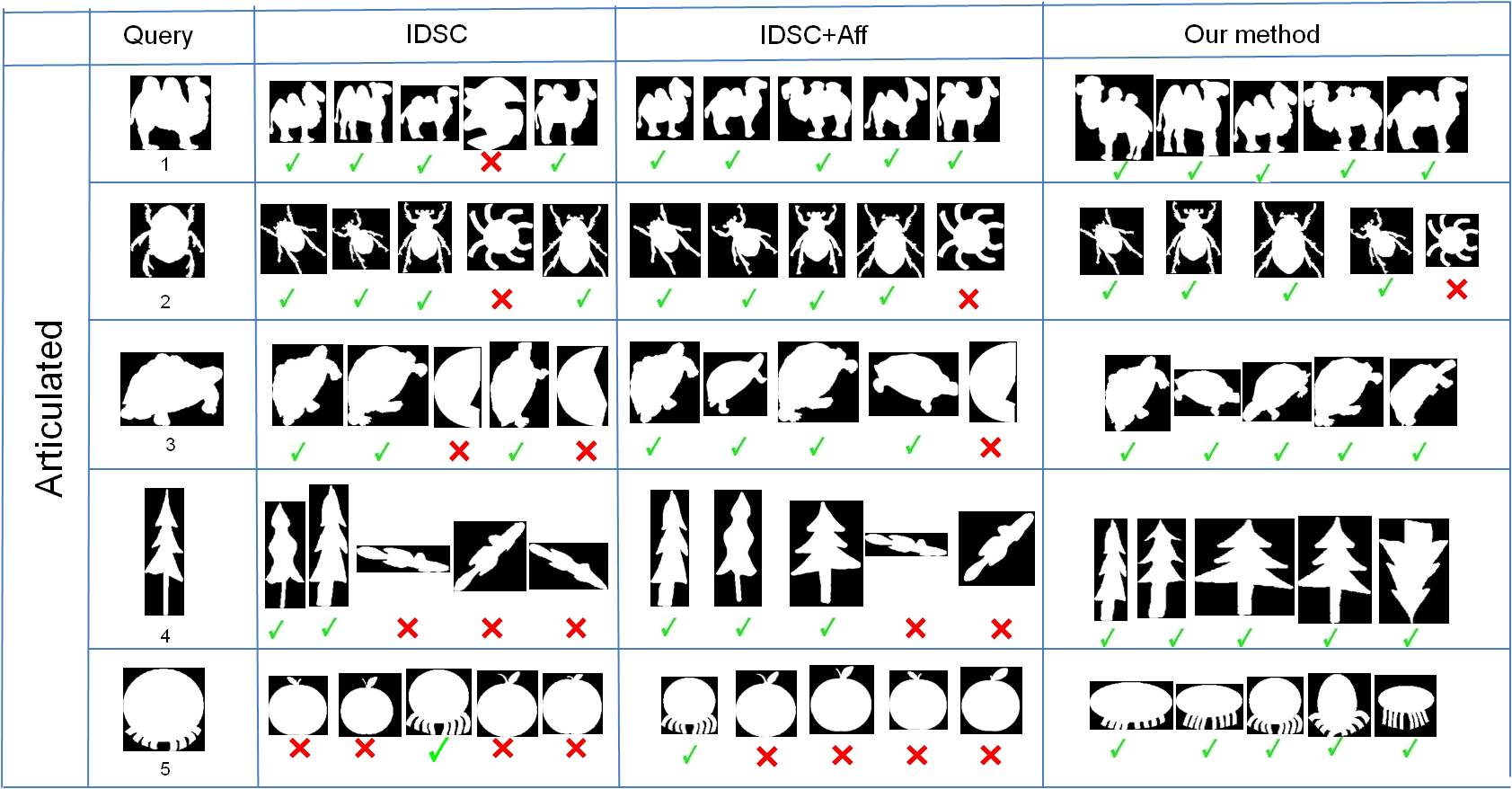}          \\
(a) The 5 most similar shapes retrieved by IDSC, IDSC+Aff and our method.  \\ \\ 
\includegraphics[scale=0.58]{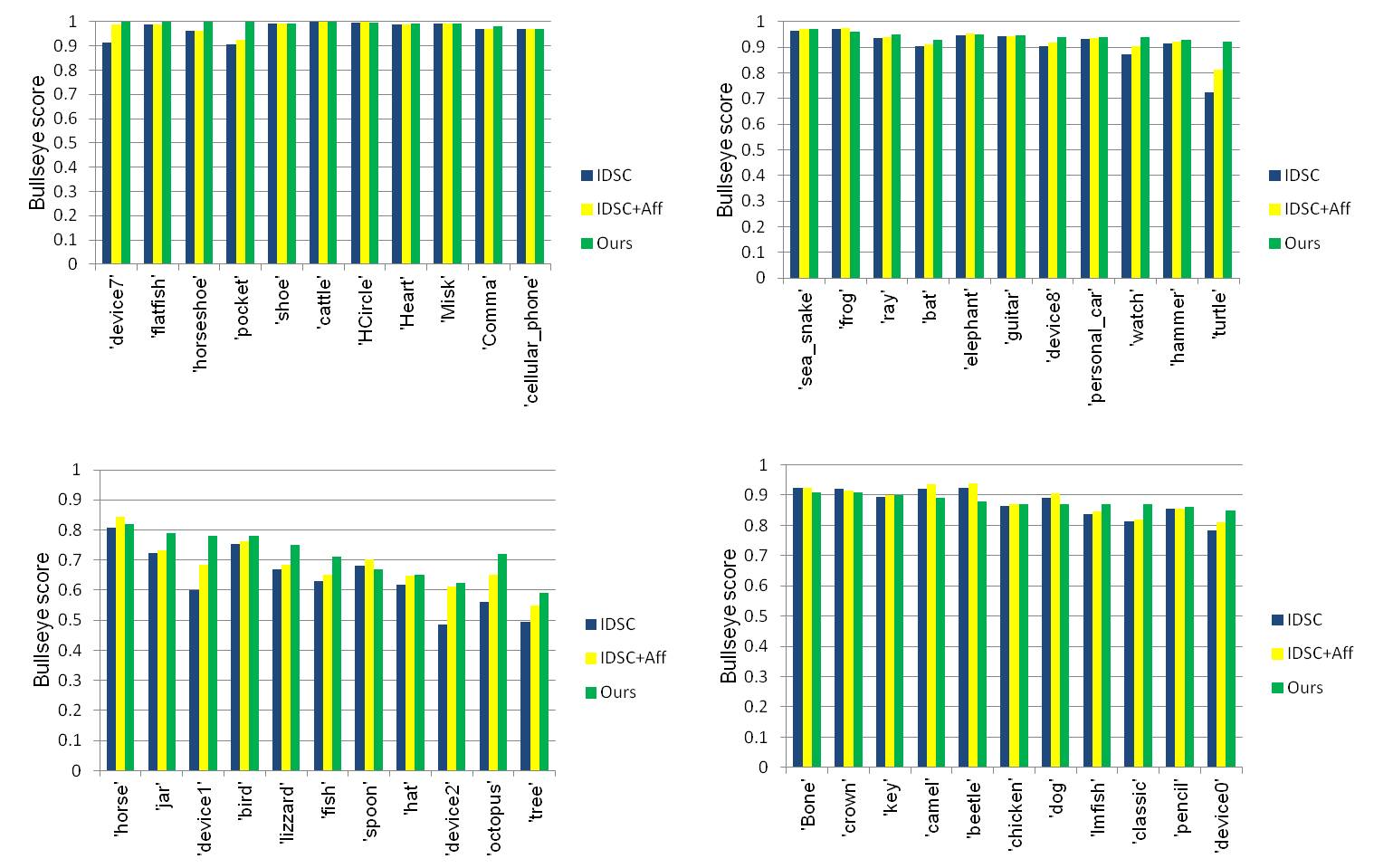}  \\
(b) Shape-wise average Bullseye scores. \\ Average: IDSC: 84.8$\%$, IDSC+Aff : 86.8$\%$, Ours: 88.3$\%$. \\
\end{tabular}      
\caption{Performance of IDSC vs. IDSC+Aff vs. our approach on articulated shapes of the basic MPEG-7 dataset which contains deformations, 
articulations, and GS-wise affine changes. }
\label{Articulated}
\end{figure*}

Figure \ref{Articulated}(a) visually demonstrates the retrieval results, while
Figure \ref{Articulated}(b) shows the average Bullseye score for the shapes that contain deformations, 
pose variations, and GS-wise affine changes.  It can be seen that IDSC+Aff and our method significantly outperform IDSC 
by $5$ to $20$ percent in the average Bullseye score on some of the shapes such as `turtle',`octopus, `pocket', `lizard', `device0', 
`device1', `device2', `device7', `classic', `horseshoe' and `tree'.  The main reason for the improvement seems to be the 
ability of these algorithms to allow the shapes to undergo different GS-wise 
affine changes.  While articulations can be handled by IDSC itself, it fails to handle these affine variations in the GSs and 
so scores low on many categories of shapes that have these variations.  Our method and IDSC+Aff 
perform almost the same since both are able to handle such variations.  Furthermore, since most of the 
shapes in the MPEG-7 dataset appear to be in this category, it is not surprising that IDSC+Aff reports good numbers for the entire dataset.

\begin{figure*}
 \centering  
\begin{tabular}{cc}
\includegraphics[scale=0.4]{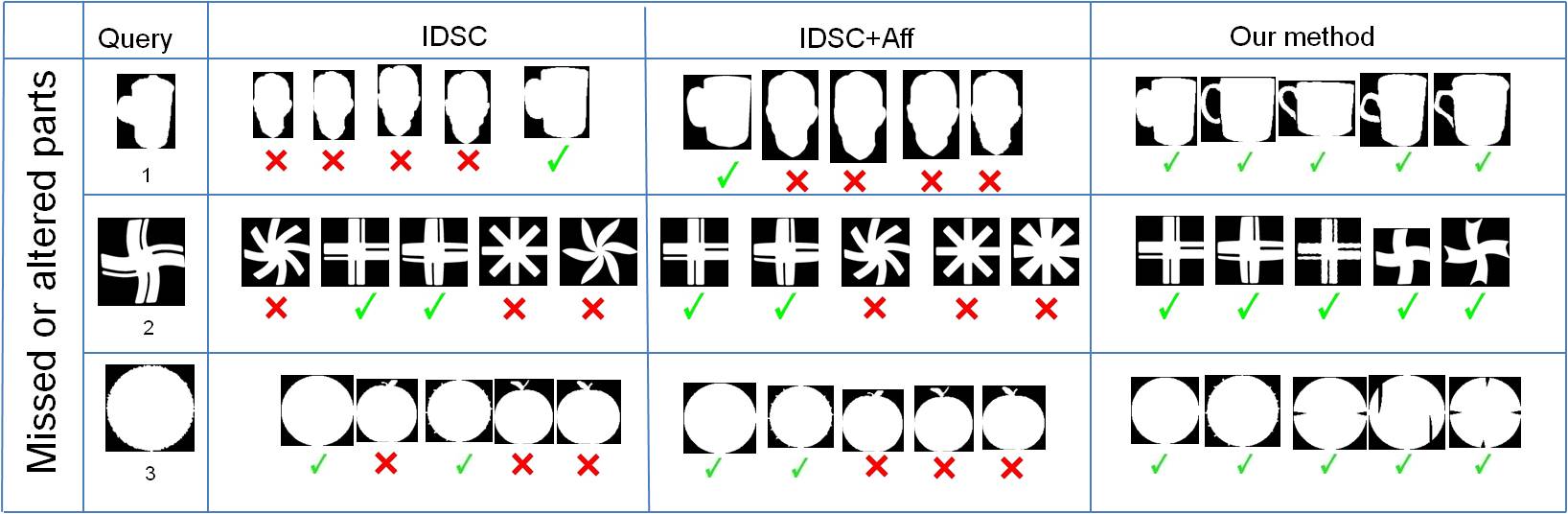}          \\
(a) The 5 most similar shapes retrieved by IDSC, IDSC+Aff and our method.  \\ \\ 
\includegraphics[scale=0.7]{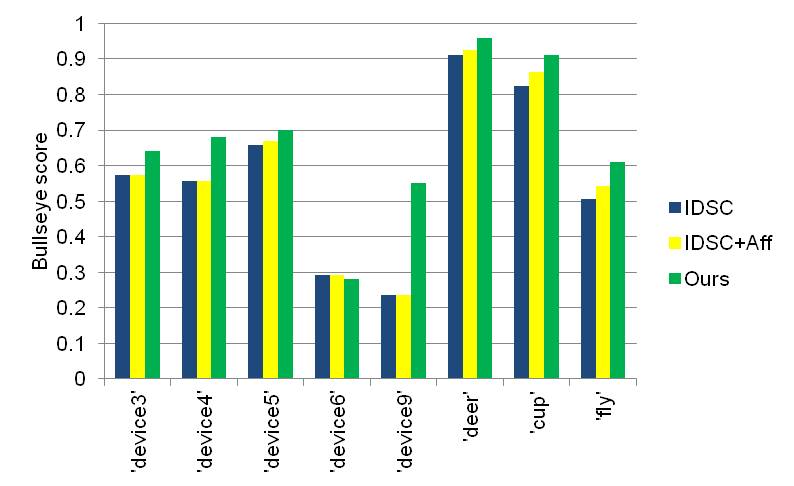}  \\
(b) Shape-wise average Bullseye scores.  \\ Average: IDSC: 57.03$\%$, IDSC+Aff: 58.28$\%$, Ours: 66.63$\%$.
\end{tabular}
\caption{Performance of IDSC vs. IDSC+Aff vs. our approach on shapes having missed or altered contour portions in the basic MPEG-7 dataset.  
}
\label{altred_parts}
\end{figure*}

The retrieval performance for shapes with missed or altered contour portions is shown visually 
in Figure \ref{altred_parts}(a) and quantitatively in Figure \ref{altred_parts}(b) where one can note that 
the improvement in the performance of our method is nearly $10$ percent compared to that of both IDSC and IDSC+Aff.  
Even affine correction of the parts seems to give very little improvement for these categories as the main challenge 
seems to be the missing portions and hence partial matching of the shapes with skips is required as opposed to a global 
matching utilized by both IDSC and IDSC+Aff.  Examples of such shape alternations include the handle of the `cup' and 
the horns of the `deer' (Fig. \ref{altred_parts}(a)).  Furthermore, global methods may miss important local differences, 
such as between `cup' and `faces', or `device-9' and `apples' as shown in Figure \ref{altred_parts}(a), due to which they 
confuse between these shapes while more local GS-based matching as is utilized in our work, is able to do much better in such circumstances.

A study of the MPEG-7 dataset shows that $90$ percent of the shapes in this dataset belong to the first two categories 
where global matching methods can perform well, especially if they handle articulations and some GS-wise normalizations 
and corrections.  Thus, it is not surprising that many of these methods (\cite{FPS,ATemlyakovCVPR2010,XChunjingPAMI2009}) 
report good overall performance numbers for this dataset.  However, as we have seen for IDSC and IDSC+Aff, 
their performance is probably not as good on the shapes in the third category, for which more advanced adaptive
matching methods are needed.  Indeed, the comparative numbers for our algorithm would have been much better if 
the dataset had more shapes in the third category.  The method proposed by Bronstein et al. \cite{ABronsteinIJCV2008}, while being 
an interesting solution to the problem with good results reported for such cases, is unfortunately computationally 
extremely expensive and testing on large datasets such as MPEG-7 is prohibitively slow.  
Thus, our method with much lower running times due to the usage of appropriate approximations and the 
resulting Dynamic Programming solution seems to be a much better practical solution for handling 
such more complex shape variations.  

Apart from natural shape variations, automatic shape extraction techniques 
such as Background Subtraction and Image Segmentation may have errors in the contour extraction process due to which
some portions may be occluded/missing, some extra portions may be added or two shapes may merge together.  
We next simulate the effect of such errors on the MPEG-7 dataset to study the ability of contour matching algorithms 
to deal with them.

\subsubsection{Partially Occluded MPEG-7 Dataset}
In order to validate the proposed technique in the presence of errors due to occlusions, 
we modified the standard MPEG-7 dataset by randomly removing
$n$ consecutive segments from a shape, where $n = k \%$ of the total number of segments in the shape,  $k$ being randomly chosen from $5-15$.

\begin{figure}
  \centering  
  \includegraphics[scale=0.45]{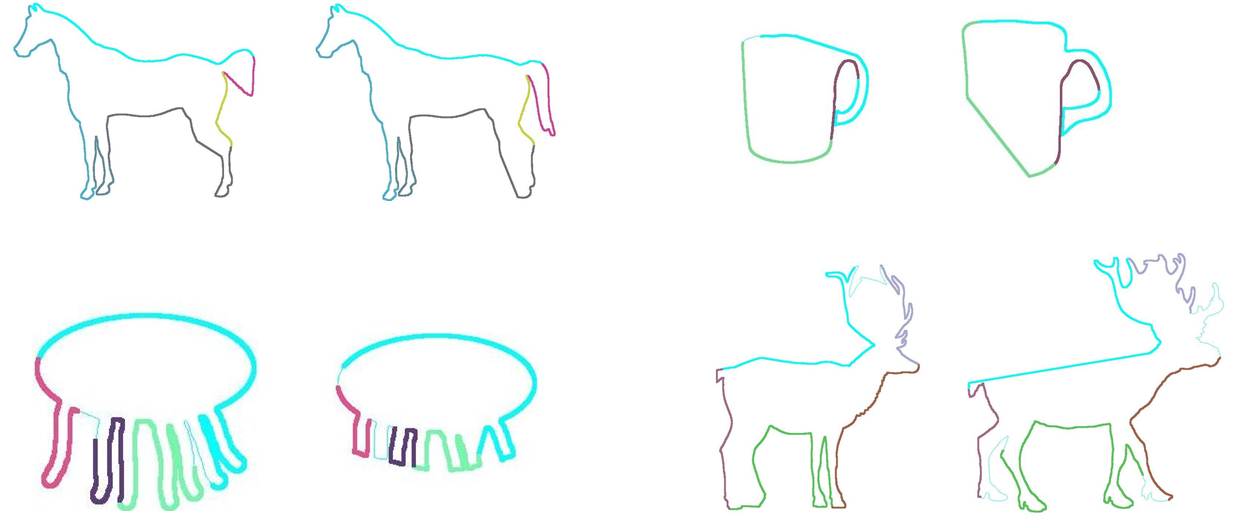}	
  \caption {Best viewed in color.  Some matching results and shape decompositions obtained between pairs of shapes that have partial occlusions.}
\label{partially_occluded_visual}
\end{figure}

\begin{figure}
  \centering  
  \includegraphics[scale=0.4]{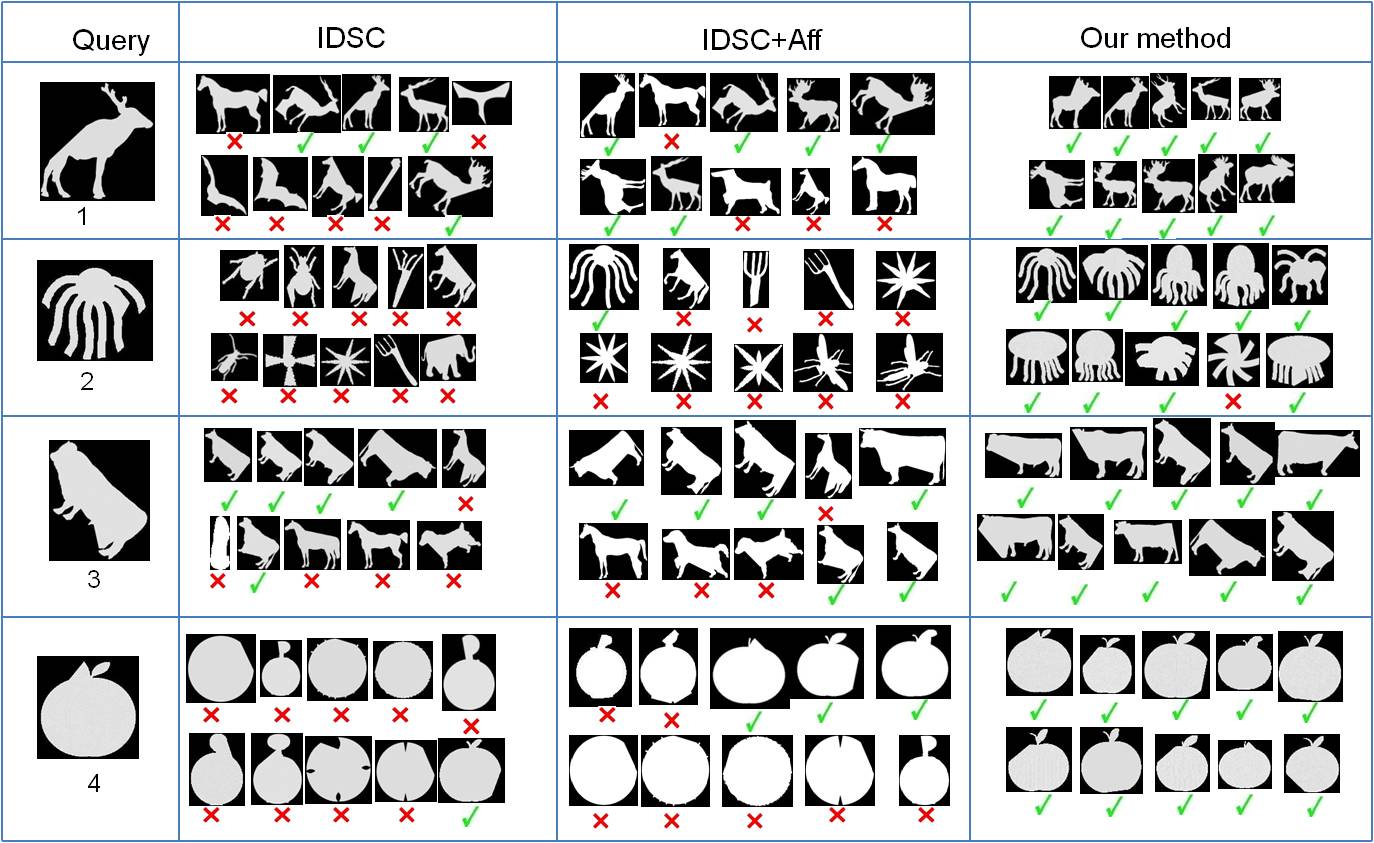}	
  \caption {The 10 most similar shapes retrieved by IDSC (second column), by IDSC+Aff (third column) and by our method (four column) for the partially occluded MPEG-7 dataset.}
\label{Retrival_res_partially_occluded}
\end{figure}

Figure \ref{partially_occluded_visual} shows some visual results of our matching algorithm on this dataset illustrating the shape decompositions, 
that may be claimed to be quite reasonable since they skip not non-matching portions while matching.  Figure \ref{Retrival_res_partially_occluded} visually 
illustrates the results for the task of shape retrieval for this dataset.  Row number 2 presents an interesting instance for the query shape 
of 'occluded octopus' where IDSC+Aff \cite{RPRC} retrieves only one shape correctly and IDSC retrieves none, 
while our method results in all but one correct retrievals.

\begin{figure}
  \centering  
  \includegraphics[scale=0.64]{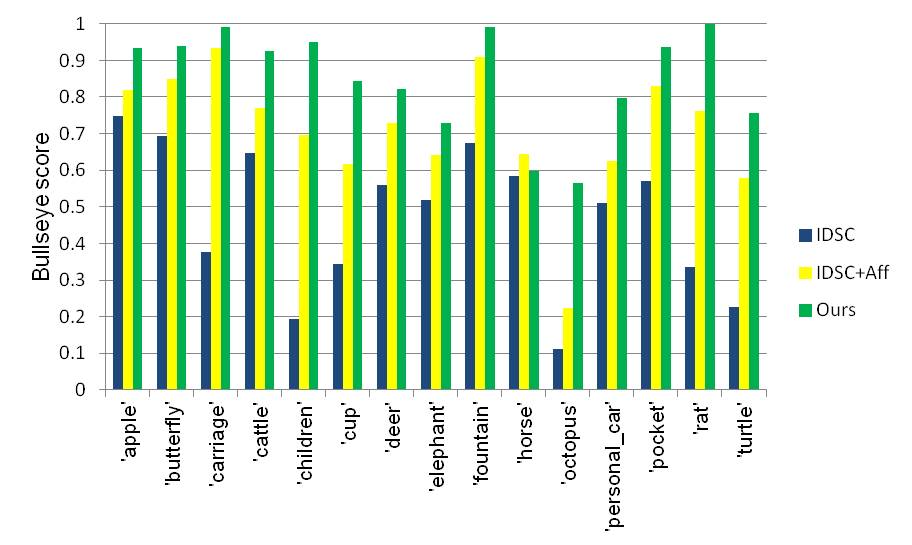}	
  \caption {Performance of IDSC vs. IDSC+Aff vs. our approach on the $15$ categories of the partially occluded MPEG-7 dataset.
 Average Bullseye score for IDSC = 47.28 $\%$, IDSC+Aff = 70.92 $\%$, and Ours = 85.2 $\%$.}
\label{Retrival_Res_POD}
\end{figure}

Additionally, we quantitatively evaluate our method on $15$ different categories of the Partially Occluded MPEG-7 dataset.  
We have chosen these $15$ categories as IDSC performs reasonably well on these categories in the original MPEG-7 dataset.  
Figure \ref{Retrival_Res_POD} illustrates the category-wise average Bullseye scores and the overall average scores.  
The performance of IDSC may be claimed to be quite unsatisfactory as the Inner Distance computation is severely affected by the 
partial occlusions present in the shapes and the global shape is also quite different.   The method proposed by \cite{RPRC} improves 
the result over IDSC but it does not explicitly model occlusions while matching and has a global approach towards shape similarity.   
On the other hand, our method performs significantly better due to its ability to model occlusions by skipping certain segments.  
This way of modeling improves the performance on an average by $38$ percent and $14$ percent improves over IDSC and the method proposed by \cite{RPRC} respectively.

\subsubsection{Merged MPEG-7 Dataset}
\begin{figure}
  \centering  
  \includegraphics[scale=0.2]{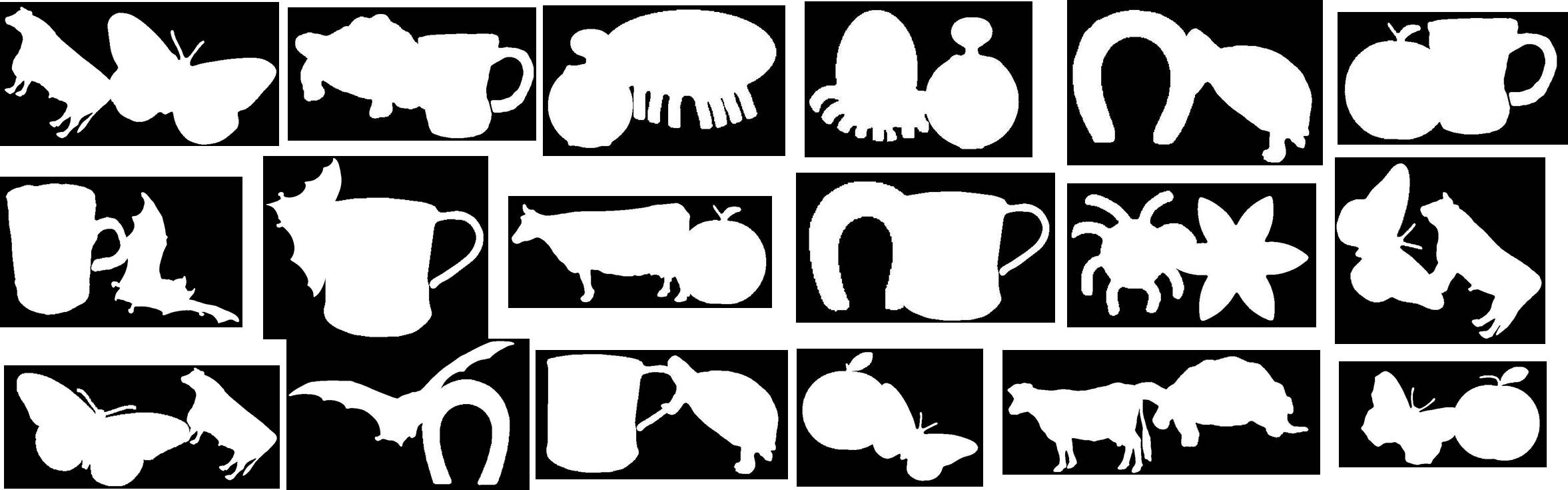}	
  \caption {Some examples from our 'Merged MPEG-7` dataset.}
\label{CombinedShapeDataSet}
\end{figure}

The other type of error that often arises in the case of contour extraction is the merging of two shapes 
with each other that can be typically seen as a result of many Background Subtraction or Image Segmentation techniques.
In order to simulate this error, we generated $100$ different combined shapes by merging two randomly chosen ones.  
Examples of such shapes are shown in Figure \ref{CombinedShapeDataSet}.

\begin{figure}
  \centering  
  \includegraphics[scale=0.24]{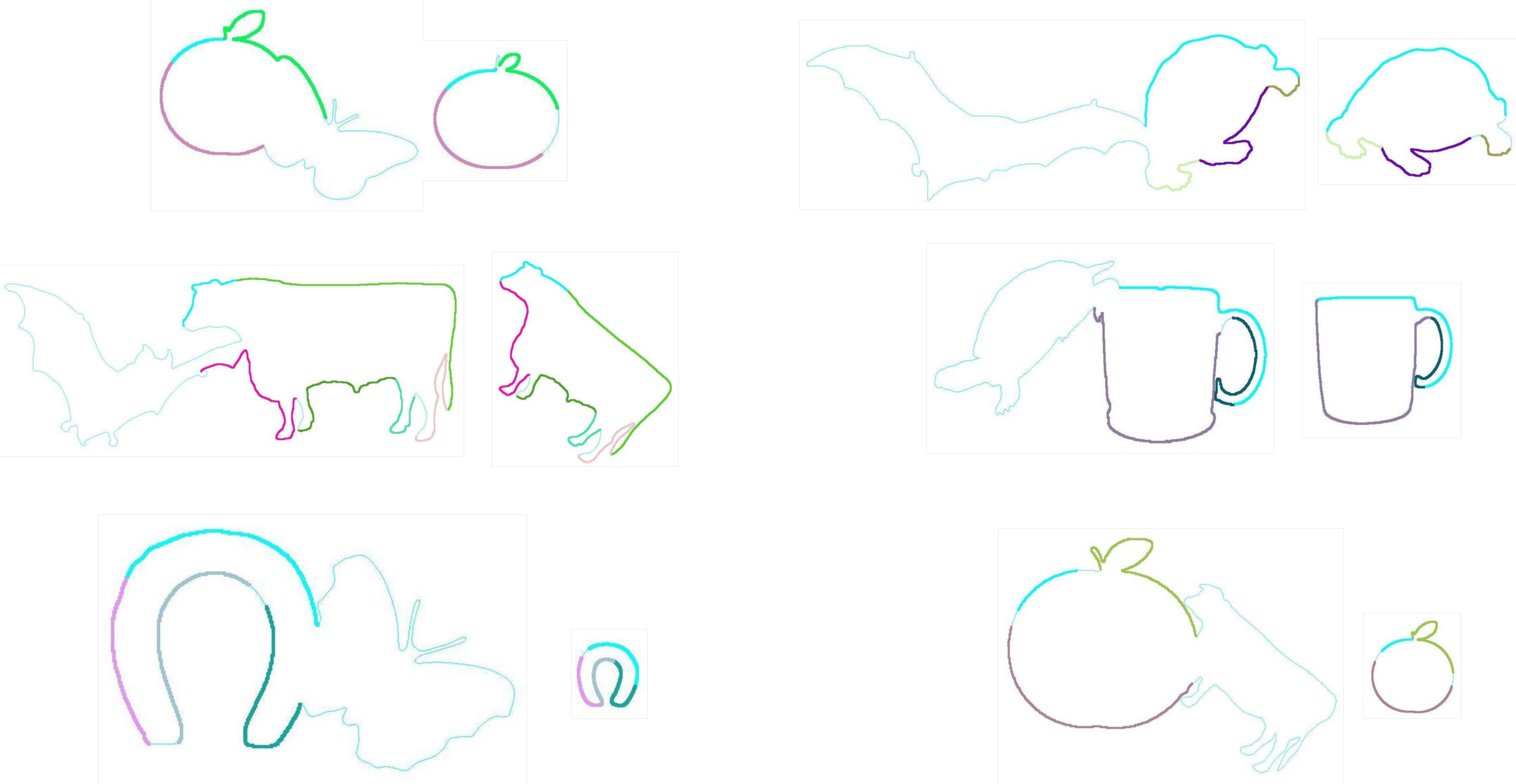}	
  \caption {Best viewed in color.  Some results of matching and extracted shape decompositions between pairs of shapes 
    when two different shapes merge into a single object shape.}
\label{Two_Object_Shapes_Merged}
\end{figure}

\begin{figure}
  \centering  
  \includegraphics[scale=0.4]{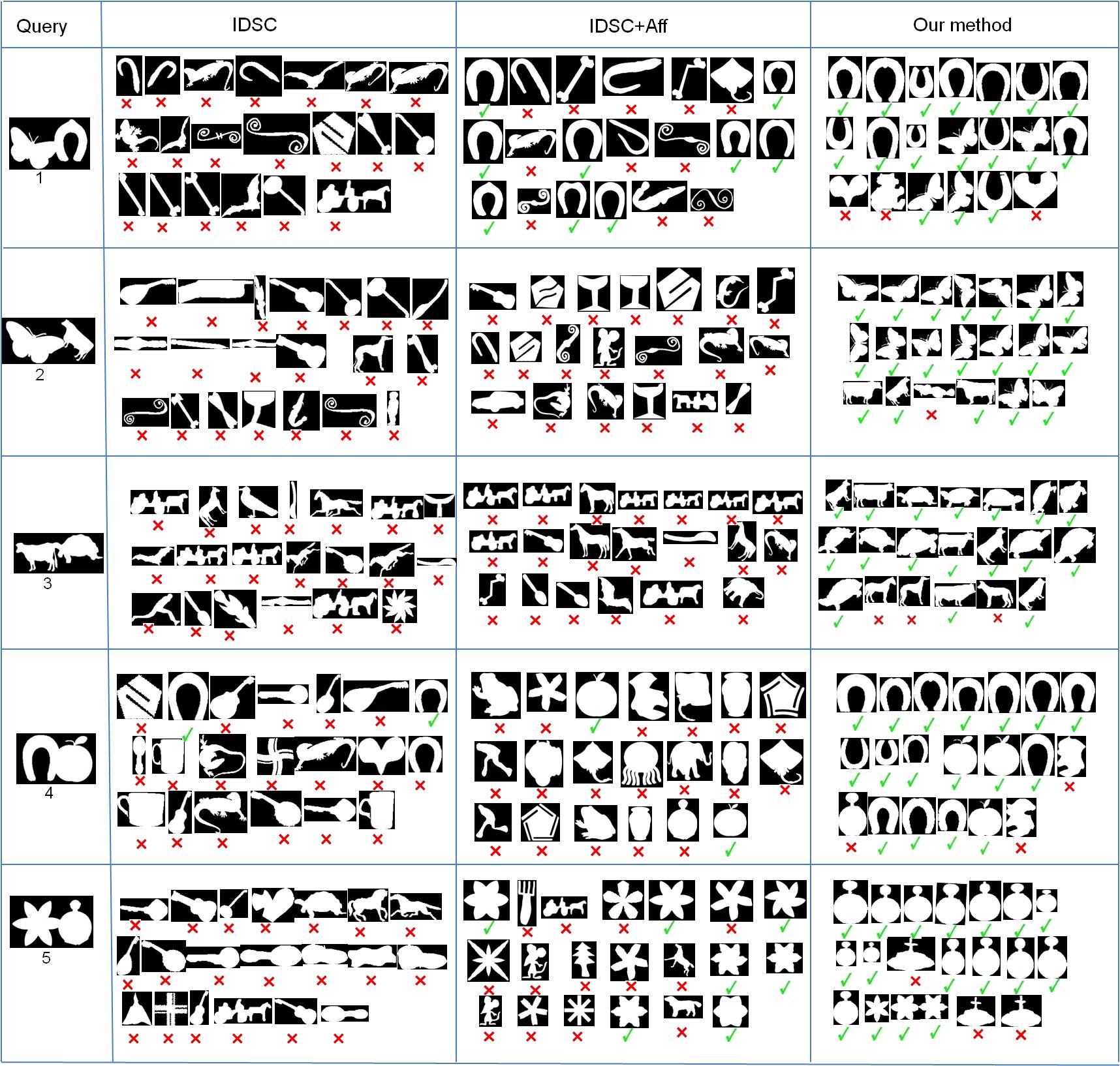}	
  \caption {The 20 most similar shapes retrieved by IDSC (second column), by IDSC+Aff (third column) and by our method (four column) for the Merged MPEG-7 dataset.}
\label{Retrival_res_combine}
\end{figure}

Figure \ref{Two_Object_Shapes_Merged} shows some matchings with their shape decompositions obtained by our algorithm
on this dataset where one can see that our method identifies the correct GS correspondences even if the queries are combined.  
 Figure \ref{Retrival_res_combine} visually compares the retrieval results obtained by our method, with those of IDSC and 
 IDSC+Aff \cite{RPRC}.  Since a query is a combined shape, the retrieved result is classified as a correct match if it contains any 
of the two shapes present in the query.  As can be seen, for most of the queries, while IDSC fails to identify even
one correct shape, and the performance of IDSC+Aff \cite{RPRC} is quite unsatisfactory, as our method retrieves mostly correct shapes.

\begin{table}
\scriptsize
\small
\renewcommand{\arraystretch}{1}
\parbox{12cm}{\caption{Comparative retrieval results on 100 different combined shapes from the MPEG-7 dataset.}\label{Combine_Shape_retrival}}
\centering
\begin{tabular}{|l|c|c|c|}\hline
   Method & Top-1 Recognition & Top-5 Recognition  & Top-10 Recognition  \\  
         &  rate (in $\%$)   & rate (in $\%$)     & rate (in $\%$)  \\            
\hline \hline 
IDSC \cite{HJD} 	 & $0.19$  & $0.146$ & $0.141$\\ \hline
IDSC+Aff \cite{RPRC} & $0.27$ & $0.234$ & $0.21$ \\ \hline 
Ours     & $0.85$  & $0.899$ & $0.843$\\ \hline
\end{tabular}
\end{table}

In addition to the visual results, Table \ref{Combine_Shape_retrival} compares the recognition rate of our method with IDSC and \cite{RPRC}
in a leave-one out environment\cite{HJD} by comparing the Top-1, Top-5 and Top-10 recognition rates for $100$ combined query 
shapes on the MPEG-7 dataset.  IDSC and the method proposed by \cite{RPRC} can be said to totally fail in this experiment while our method performs reasonably well.  
This is due to a global criteria for matching and all methods that do not explicitly handle partial matchings should fail miserably on this dataset.

\subsection{A Real Background Subtraction Dataset}
\begin{figure}
  \centering  
  \includegraphics[scale=0.42]{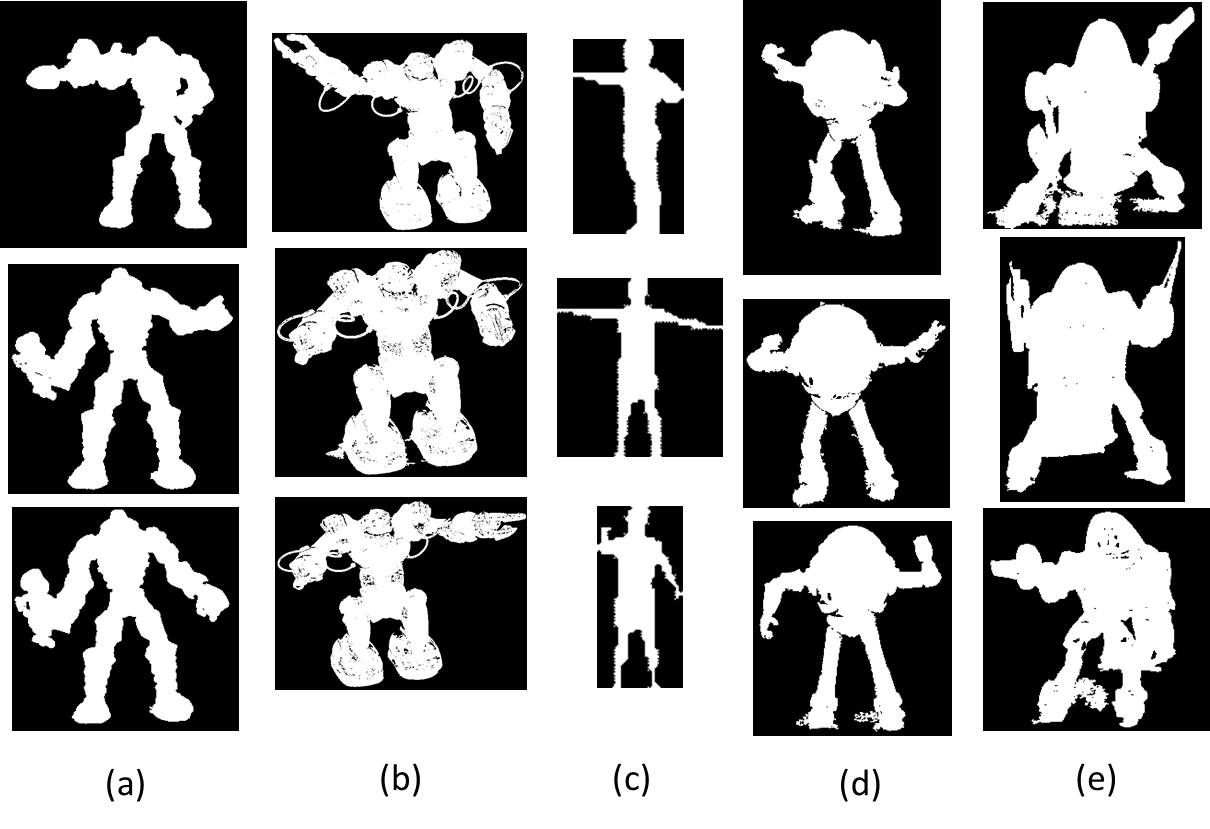}	
  \caption {Some example shapes from a real Background Subtraction dataset \cite{RPRC}. }
\label{RB_DataSet}
\end{figure}

We next evaluate our algorithm on a real Background Subtraction dataset provided by Gopalan et al. \cite{RPRC}.  
This dataset contains $50$ images of $10$ shapes per class from $5$ different classes.  The dataset has a wide 
range of non-planar articulations with significant self-occlusions and the images are captured under different 
viewpoint variations.  Many real world scenarios are characterized by such variations.
Some examples of such shapes are shown in Figure \ref{RB_DataSet}.  

Table \ref{gopalan_table} compares our retrieval results with those of IDSC \cite{HJD} and Gopalan et al. \cite{RPRC} 
in a leave-one out environment by listing the Top-1 recognition rate and the Bullseye score.  
In addition to its shortcomings already discussed, IDSC \cite{HJD} also fails to capture the 3D articulations of these shapes.
Gopalan et al. \cite{RPRC} attempt to do so by performing an affine normalization of GSs.  
However, many shape variations still remain unmodeled because of the lack of handling occlusions.
Thus, it is not surprising that our method significantly outperforms \cite{HJD} and \cite{RPRC}.

\begin{table}
\scriptsize
\small
\renewcommand{\arraystretch}{1}
\parbox{15cm}{\caption{Comparative retrieval results on a real Background Subtraction dataset.}\label{gopalan_table}}
\centering
\begin{tabular}{|l|c|c|c|c|c|}\hline
  Method & Top-1 Recognition rate (in $\%$) & Bullseye score (in $\%$)  \\  
\hline \hline 
IDSC \cite{HJD} 	 & $58$ & $39.4$ \\  \hline
IDSC + Aff \cite{RPRC}  & $80$ & $63.8$ \\ \hline
Ours     &$94$  & $81.5$ \\ \hline
\end{tabular}
\end{table}

\subsection{2D Mythological Creatures dataset}
\begin{figure}
  \centering  
  \includegraphics[scale=0.26]{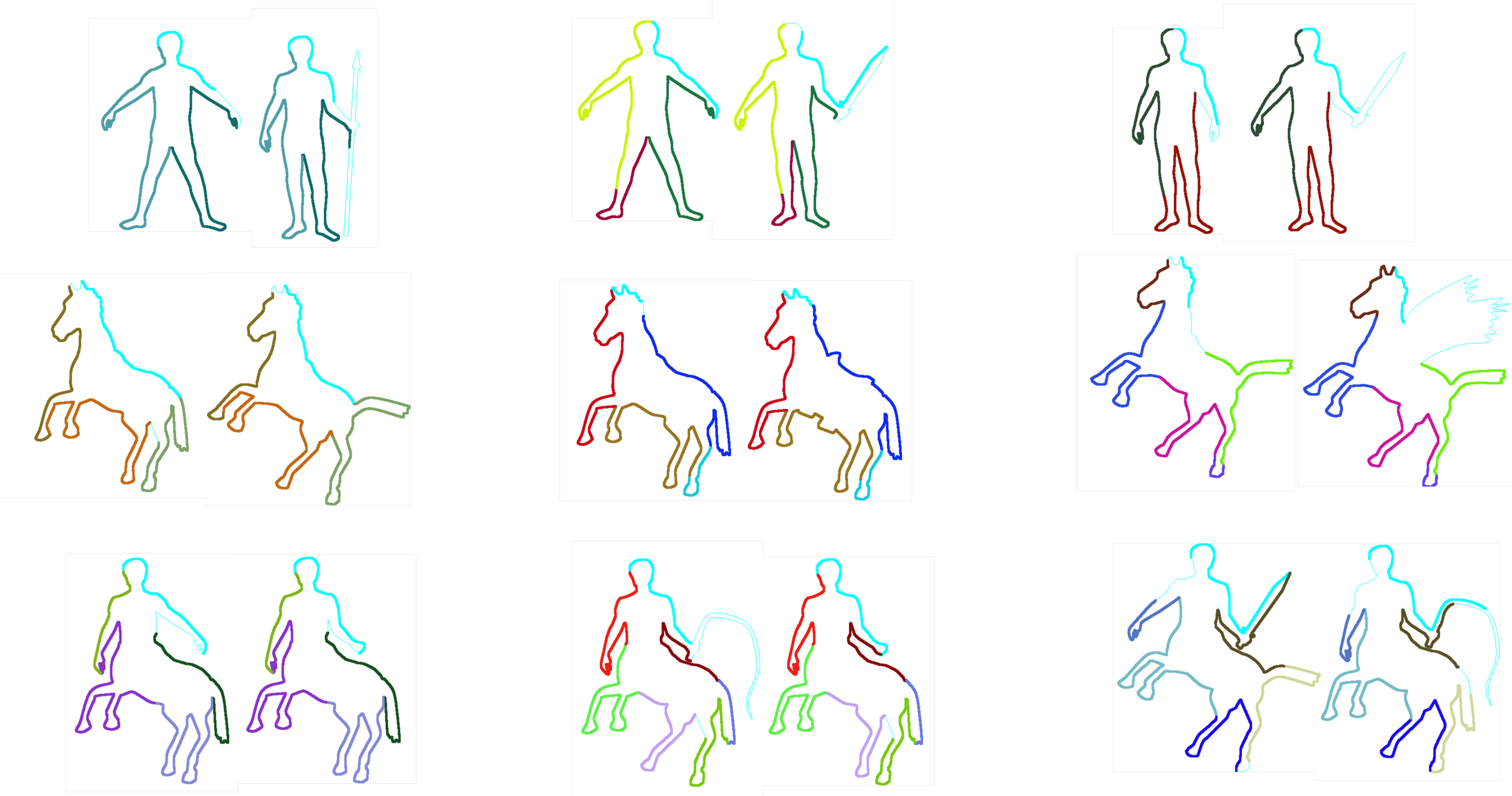}	
  \caption {Best viewed in color.  Some results of extracted shape decompositions between pairs of shapes on the 2D Mythological Creatures dataset \cite{ABronsteinIJCV2008}.
  Note that non-matching portions are skipped.}
\label{Bronstein_dataset_visual_pd}
\end{figure}

To further test the ability of our algorithm in shape decomposition and matching in the presence of occlusions,
we test it on the 2D Mythological Creatures dataset \cite{ABronsteinIJCV2008}.
This dataset contains fifteen shapes of horses, humans and centaurs, that have different articulations and partial occlusions.
Figure \ref{Bronstein_dataset_visual_pd} shows the visual results of our matching algorithm in the presence of articulations, 
deformations and occlusions.  As can be seen, we are able to identify the GS-correspondences across the shapes even in the presence of a lot of distractions.  
Thus, it may be claimed that our algorithm can be used for such contour correspondence problems.    
 
\section{Conclusion}
\label{conclude}
We have introduced an adaptive approach for shape matching that allows for a different affine variation in different portions of a shape.  The 
method does not assume a given shape decomposition {\em a priori} but determines such decomposition while matching, which makes the
matching quite robust.  Efficiency is achieved via Dynamic Programming by enforcing an ordering constraint.
Further, partial occlusions and errors in 
contour extraction are handled by allowing skips while matching.  
Experiments indicate that the method might be useful compared to existing techniques, especially in the case of partial occlusions, extra
contour portions and merged shapes that might arise in many situations, including automatic shape extraction using techniques such as 
Background Subtraction and Image Segmentation.

\bibliographystyle{elsarticle-num}
\bibliography{ref}

\end{document}